\documentclass[10pt,twocolumn,letterpaper]{article}

\usepackage[pagenumbers]{cvpr}

\renewcommand{\paragraph}[1]{\vspace{.5em}\noindent\textbf{#1}} %

\usepackage{colortbl}
\usepackage{multirow} %
\usepackage{tabularx} %

\usepackage[most]{tcolorbox}
\newtcolorbox{promptbox}[1]{enhanced, colback=white, colframe=black!35, boxrule=0.4pt, arc=1pt,
  left=6pt, right=6pt, top=4pt, bottom=4pt, title={#1}, fonttitle=\small\bfseries, coltitle=black,
  colbacktitle=black!6, toptitle=2pt, bottomtitle=2pt, fontupper=\scriptsize,
  before upper={\setlength{\parskip}{2pt}}}
\newcommand{\promptlabel}[1]{\noindent{\sffamily\scriptsize\bfseries\color{black!55}\MakeUppercase{#1}}\par}

\newcommand{\dataset}{DUALingo\xspace}
\newcommand{\numObjectsExact}{70{,}689\xspace}
\newcommand{\numTasksExact}{270{,}884\xspace}

\newcommand{\numObjects}{71K\xspace}
\newcommand{\numTasks}{271K\xspace}
\newcommand{\numRoles}{542K\xspace}

\newcommand{\numPoints}{16{,}384\xspace}

\newcommand{\shareObjectLevelTasks}{38.74\%\xspace}
\newcommand{\shareComponentLevelTasks}{39.30\%\xspace}
\newcommand{\shareFunctionalPoseTasks}{21.95\%\xspace}

\newcommand{\numHumanTasks}{1{,}000\xspace}

\newcommand{\numTrainTasks}{254{,}035\xspace}

\newcommand{\numTrainObjects}{66{,}456\xspace}

\newcommand{\shareRemovedBad}{12.6\%\xspace}

\definecolor{cvprblue}{rgb}{0.21,0.49,0.74}
\usepackage[pagebackref,breaklinks,colorlinks,allcolors=cvprblue]{hyperref}

\def\confName{CVPR}
\def\confYear{2027}

\title{
DUALingo: A Dual-Hand Instruction-Guided Dataset for Affordance Grounding on 3D Objects
}

\author{Ming-Lun Lee$^{1}$ \quad Yi-En Li$^{2}$ \quad Feng-Dian Su$^{2}$ \quad Ming-Han Lee$^{2}$\\
Kun-Ru Wu$^{2}$ \quad Krzysztof Czarnecki$^{1}$ \quad Yu-Chee Tseng$^{2}$\\[0.5em]
$^{1}$University of Waterloo \quad $^{2}$National Yang Ming Chiao Tung University\\[0.3em]
{\tt\small \{m392lee, k2czarne\}@uwaterloo.ca}\\
{\tt\small \{ieen.cs13, sufengdian.cs14, mhlee.cs09, wufish, yctseng\}@nycu.edu.tw}}

\begin{document}
\twocolumn[{
\renewcommand\twocolumn[1][]{#1}
\maketitle
\begin{center}
\centering
\includegraphics[width=\textwidth]{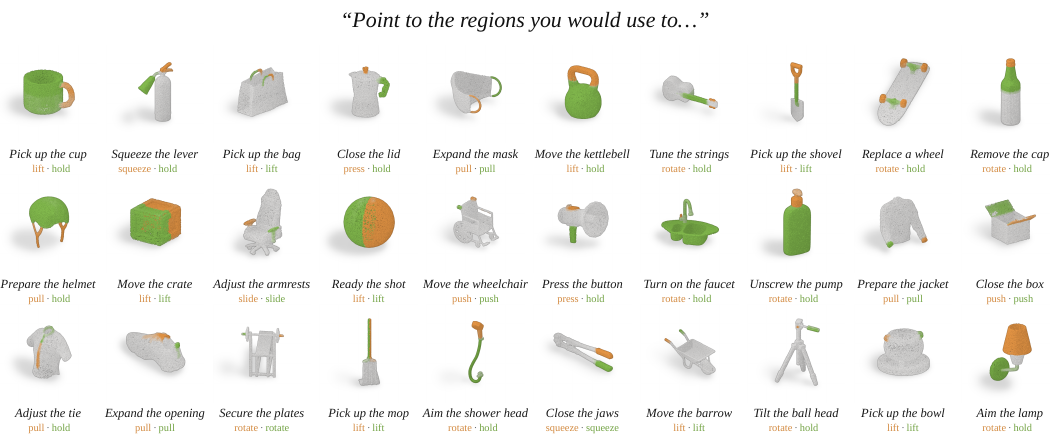}
\captionof{figure}{\textbf{Examples from \dataset.} Each object-task pair specifies two cooperative, hand-agnostic roles grounded as 3D surface heatmaps, illustrating diverse objects, task granularities, and bimanual coordination patterns.}
\label{fig:teaser}
\end{center}
}]

\begin{abstract}
Affordance grounding localizes where an agent should interact with an object to perform a task.
Many everyday manipulation tasks, however, require \textbf{bimanual affordance}, in which two hands assume complementary and cooperative roles to accomplish a shared goal.
Existing 3D affordance datasets primarily ground single interaction regions, and manually annotating dual-hand, cooperative contacts does not scale.
To address this gap, we introduce \textbf{DUALingo}, a large-scale dataset for task-conditioned, role-aware bimanual 3D affordance grounding on diverse everyday objects.
DUALingo represents each task as two cooperative, hand-agnostic roles, each grounded as a heatmap on the object surface.
It contains 271K free-form tasks on 71K objects, generated by a fully automated pipeline of open-weight foundation models without human labeling.
Expert ratings and a validated LLM judge confirm the quality of the resulting annotations.
Trained on DUALingo, existing 3D affordance models learn to ground both cooperative roles from natural-language instructions, with the best-performing model achieving a \textbf{more than threefold improvement in aIoU over the zero-shot baseline}, from 10.3 to 37.8.
Robotic manipulation demonstrations illustrate the actionability of the predicted bimanual affordances in performing cooperative tasks.
These results position DUALingo as a scalable foundation for learning everyday bimanual affordances.
Project page: \mbox{\href{https://nycu-pair-labs.github.io/DUALingo/}{\texttt{nycu-pair-labs.github.io/DUALingo}}}.
\end{abstract}

\section{Introduction}
\label{sec:introduction}

Embodied agents operating in the physical world must understand where and how
to interact with surrounding objects. \emph{Visual affordance
learning}~\cite{myers2015affordance} models
the possible interactions between an agent and its environment~\cite{gibson1979ecological,montesano2008learning}. Given an image or
a 3D object and an instruction, \emph{affordance
grounding}~\cite{do2018affordancenet} localizes the pixels or surface points
required for the intended interaction, such as the handle for opening a
drawer. As a bridge between perception and action, affordance grounding
supports robotic manipulation~\cite{geng2023rlafford,moldovan2012relational,bjorck2025groot},
warehouse automation~\cite{li2023package}, and human--robot
collaboration~\cite{li2025preciseaffordances}.

Many everyday tasks, however, require two hands that play distinct but
cooperative roles~\cite{guiard1987asymmetric,rakita2019shared}, such as
rotating a bottle cap against counter-torque at the
body~\cite{lin2024twisting} or carrying a storage bin by its two
handles~\cite{jin2026umibench}. This work studies \textbf{bimanual affordance grounding}~\cite{zhao2023dualafford,heidinger2025twohands}: given a 3D object and a free-form task, the goal is to infer two
cooperative roles and ground each role as a heatmap on the object surface.
Compared with single-region affordance grounding, bimanual affordance grounding introduces two additional challenges. First, the two contact regions are \textbf{interdependent}, as the role of one hand constrains where the other must act~\cite{pais2018constraints,grannen2023stabilize,plonka2024spatial}. Second, the cooperative roles are \textbf{task-dependent}: opening a box requires the hands to pull in opposite directions~\cite{zhou2025yoto}, whereas lifting the same box requires them to act together.

3D affordance learning has progressed rapidly, from category-level
labels~\cite{deng2021affordancenet,xu2022partafford} to open-vocabulary,
automatically annotated, and video-based
grounding~\cite{nguyen2023openad,li2024laso,yu2024seqafford,lee2025affogato,zhu2025agpil,wang2026videoafford,mao2026vagnet}.
However, these datasets primarily ground one region per query, leaving the
relation between two contacts implicit. Existing bimanual datasets capture two-hand interaction through motion capture on a limited set of objects~\cite{fan2023arctic,zhan2024oakink2,liu2024taco,fu2025gigahands},
dexterity-rated grasp pairs~\cite{zhai2022da2}, or hand masks in 2D
images~\cite{heidinger2025twohands}.
However, manually annotating task-conditioned, dual-hand roles at scale is costly. A comparison of our dataset with existing ones is provided in \cref{tab:dataset-comparison}.

In this work, we introduce \textbf{\dataset}, a large-scale, instruction-guided dataset for role-aware bimanual 3D affordance grounding (\cref{fig:teaser}). It comprises \numTasks tasks across \numObjects objects, with \numRoles role-specific heatmaps covering object-, component-, and function-level natural-language instructions.
Unlike prior datasets that primarily annotate where interactions occur, \dataset additionally specifies the cooperative role of each contact region in completing a task. Each free-form task is decomposed into two hand-agnostic roles, expressed as action primitives such as \texttt{hold} and \texttt{rotate}, each grounded as a heatmap on the object surface.

To build \dataset at scale, we design a \textbf{fully automated annotation pipeline}
that orchestrates open-weight foundation
models~\cite{gemmateam2026gemma4,clark2026molmo2,ravi2024sam2}, without human
labeling. A multimodal LLM first identifies the manipulable components of each
object, proposes bimanual tasks, and decomposes each task into two cooperative
roles through a physical operation plan. A pointing model then localizes each
role across views with adaptive multi-point grounding, which captures repeated targets and improves region coverage. Role-aware segmentation converts these contacts
into masks with reduced cross-role leakage. Finally, confidence-aware
aggregation fuses the views into a pair of 3D heatmaps. Expert ratings confirm the quality of the resulting annotations.

\begin{table}[t]
\centering
\caption{\textbf{Comparison with existing affordance datasets.} DUALingo uniquely combines 3D grounding, free-form tasks, bimanual interactions, and hand-agnostic action roles (Role-aware).}
\label{tab:dataset-comparison}
\vspace{-6pt}
\begingroup
\small
\setlength{\tabcolsep}{2.9pt} %
\renewcommand{\arraystretch}{1.08}
\begin{tabular}{lcccc} %
\toprule
Dataset & 3D & Free-form & Bimanual & Role-aware \\
\midrule
3D AffordanceNet~\cite{deng2021affordancenet} & \checkmark & -- & -- & -- \\
LASO~\cite{li2024laso} & \checkmark & \checkmark & -- & -- \\
Affogato~\cite{lee2025affogato} & \checkmark & \checkmark & -- & -- \\
DA$^2$~\cite{zhai2022da2} & \checkmark & -- & \checkmark & -- \\
DualAfford~\cite{zhao2023dualafford} & \checkmark & -- & \checkmark & -- \\
2HANDS~\cite{heidinger2025twohands} & -- & \checkmark & \checkmark & -- \\
DHAGrasp~\cite{li2025dhagrasp} & \checkmark & -- & \checkmark & -- \\
BiPreManip~\cite{shen2026bipremanip} & \checkmark & -- & \checkmark & -- \\
\rowcolor{gray!12}
\textbf{\dataset{} (Ours)} & \checkmark & \checkmark & \checkmark & \checkmark \\
\bottomrule
\end{tabular}
\endgroup
\vspace{-5pt}
\end{table}

Alongside the dataset, we benchmark three 3D affordance grounding
models~\cite{nguyen2023openad,li2024laso,lee2025affogato}, adapted to predict dual-hand roles and their heatmaps from the task. Without training on
\dataset, the original models largely localize the same region for both
roles, with AUC near the chance level of 50. Trained on \dataset, all three models improve aIoU more than threefold. In particular, the adapted Espresso-3D~\cite{lee2025affogato} improves from 10.3 to \textbf{37.8 aIoU} and from 53.7 to \textbf{79.5 AUC}. However, role accuracy remains below 80\%, leaving substantial room for improvement.
We deploy the trained model in
ManiSkill3~\cite{tao2025maniskill3} on objects from sources outside \dataset.
The predicted heatmaps serve as contact targets for a dual-arm robot, enabling
coordinated manipulation across multi-granular tasks.
\dataset thus provides a foundation for advancing research on bimanual affordance reasoning, hand–object interaction, and cooperative robotic manipulation.

To summarize, our contributions are as follows:
\begin{itemize}
    \item We introduce \dataset, a large-scale dataset for task-conditioned,
    role-aware bimanual 3D affordance grounding with \numTasks tasks on
    \numObjects objects. To the best of our knowledge, it is the first 3D
    dataset that pairs free-form tasks with cooperative, hand-agnostic roles
    grounded as surface heatmaps.
    \item We benchmark representative 3D affordance grounding models and
    find that single-region models transfer poorly to the bimanual
    setting. Training on our data improves every model on all metrics, while
    inferring the cooperative roles of a task remains open.
    \item We demonstrate that the predicted affordances guide
    coordinated bimanual manipulation across multi-granular tasks, laying a
    foundation for cooperative robots.
\end{itemize}

\section{Related Work}
\label{sec:related}

\paragraph{Visual affordance grounding.}
\leavevmode Visual affordance grounding localizes the object regions that support a
described interaction. Early methods learn affordances from pixel-level labels and masks on 2D images~\cite{myers2015affordance,do2018affordancenet}.
3D AffordanceNet~\cite{deng2021affordancenet} establishes a benchmark of
point-wise 3D affordance heatmaps, and PartAfford~\cite{xu2022partafford} grounds
affordances on functional parts. OpenAD~\cite{nguyen2023openad} and
LASO~\cite{li2024laso} extend 3D grounding to unseen affordance labels and
natural-language questions. More recent work handles complex
instructions~\cite{chu2025affordancellm}, sequential
tasks~\cite{yu2024seqafford}, and multimodal
observations~\cite{zhu2025agpil}. 
Despite broader vocabularies and input modalities, these works primarily address single-hand interactions. In contrast, \dataset annotates two cooperative, role-specific contact regions for each task.

\paragraph{Bimanual manipulation datasets.}
Bimanual manipulation datasets capture two-hand interaction through different
signals. DA$^2$~\cite{zhai2022da2} generates dexterity-rated dual-arm grasp pairs for
thousands of objects, and DualAfford~\cite{zhao2023dualafford}, BiAssemble~\cite{shen2025biassemble}, A3D~\cite{liang2026a3d}, BiPreManip~\cite{shen2026bipremanip}, DHAGrasp~\cite{li2025dhagrasp}, and BiDexAffordance~\cite{liu2026bidexaffordance} learn dual-gripper and dual-hand affordances for bimanual tasks. Recent work further predicts such affordances with VLM-guided reasoning~\cite{hahne2026taskaware}. Among motion-capture datasets, ARCTIC~\cite{fan2023arctic} focuses
on articulated objects, OakInk2~\cite{zhan2024oakink2} decomposes complex
tasks into interdependent primitives, TACO~\cite{liu2024taco} organizes
tool--action--object compositions, and GigaHands~\cite{fu2025gigahands} pairs
bimanual motion with language. Closest to our setting, 2HANDS~\cite{heidinger2025twohands} extracts
task-conditioned left- and right-hand affordance masks from human activity
videos. These works provide rich supervision for reconstruction, synthesis,
2D grounding, and dual-arm manipulation. \dataset complements them by grounding task-conditioned,
hand-agnostic roles on 3D objects.

\paragraph{Foundation model-based affordance annotation.}
Foundation models now scale affordance supervision without the human
demonstrations that video-based methods
require~\cite{bahl2023vrb,wang2026videoafford,mao2026vagnet}.
LASO~\cite{li2024laso} and SeqAfford~\cite{yu2024seqafford} diversify
instructions with GPT-4 and GPT-4o~\cite{openai2023gpt4,openai2024gpt4o},
and Affogato~\cite{lee2025affogato} automates both query generation and 3D
heatmap construction. Other work directly elicits (M)LLM affordance priors
for training-free grounding~\cite{zhang2025a4agent}, part and
attribute knowledge~\cite{rai2024strategies}, or heatmap
decoding~\cite{qian2024affordancellm}. Part-level cues further anchor
instructions to the sub-structures they refer
to~\cite{xu2022partafford,wan2025instructpart}. Our work extends this paradigm
from single-region affordances to the two cooperative roles of a bimanual
task.

\section{\dataset Dataset}
\label{sec:dataset}

\dataset is a large-scale dataset of everyday 3D objects, providing free-form bimanual task instructions, their corresponding cooperative roles, and role-specific surface heatmaps for each object. Given a natural-language task instruction, \dataset supports the training of embodied agents to identify precise contact regions on 3D objects for coordinated bimanual manipulation.

\begin{figure*}[t]
\centering
\includegraphics[
    width=\textwidth
]{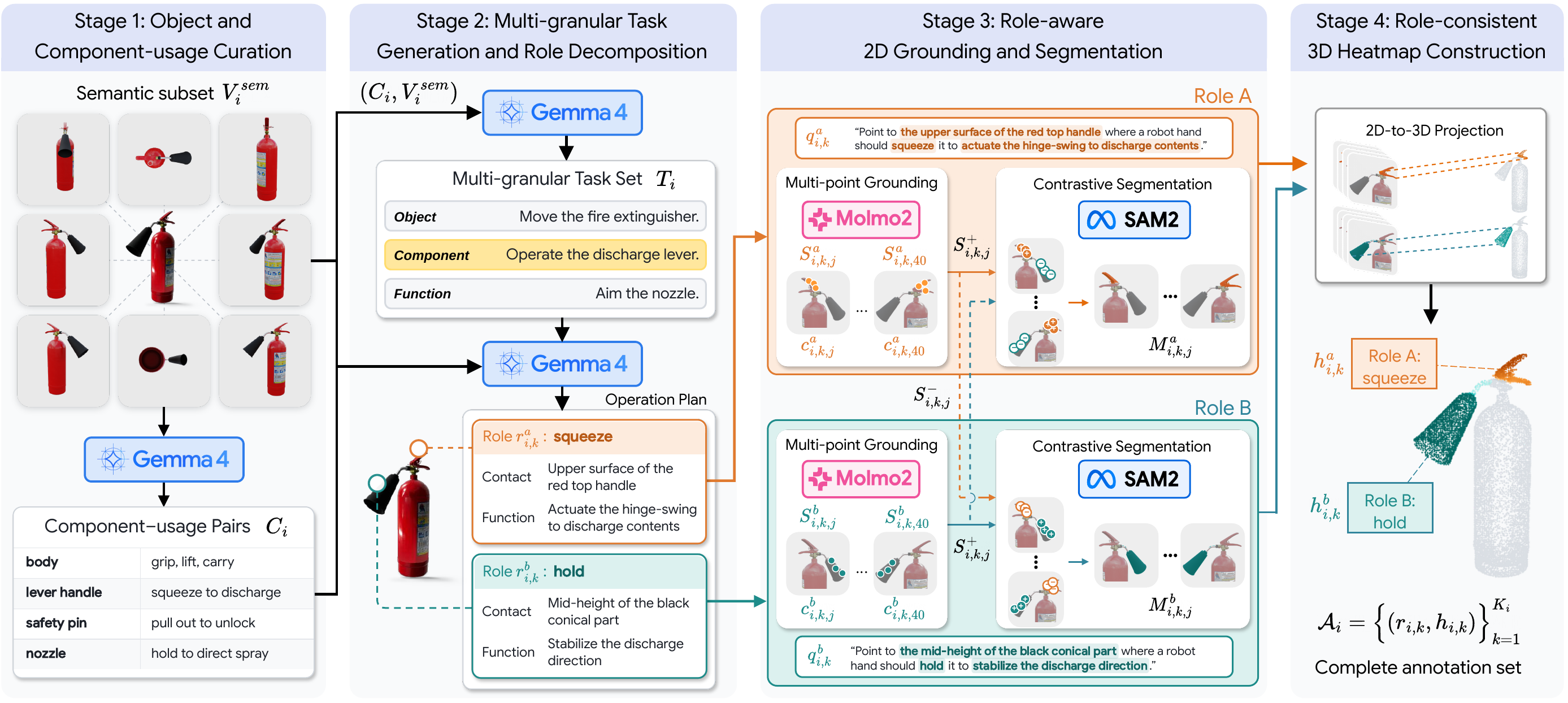}
\caption{\textbf{The \dataset annotation pipeline.} Given the rendered views of
a 3D object, foundation models curate its manipulable components, generate
multi-granular bimanual tasks, and decompose each task into two cooperative
roles. Both roles are then grounded and segmented in the 2D views and
projected onto the point cloud as a pair of 3D heatmaps.}
\label{fig:pipeline}
\end{figure*}

\subsection{Source Assets}
\label{sec:source-assets-main}

\leavevmode \dataset contains multimodal data: 3D point clouds, multi-view RGB-D observations, and textual tasks for 3D grounding. It is primarily curated from three data sources. 
\begin{itemize}
\item 
\textbf{Objaverse}~\cite{deitke2023objaverse} contains over 800K textured 3D models collected from diverse online repositories. It covers a broad range of object categories and provides rich 3D assets for research in 3D vision and embodied AI.
\item 
\textbf{G-Objaverse}~\cite{zuo2024sparse3d}, built upon~\cite{deitke2023objaverse}, provides 40 multi-view RGB-D observations for each 3D object, together with the corresponding camera parameters. These observations support multi-view visual understanding and 2D contact localization, while the depth and camera information enables projection of localized regions into 3D space.
\item 
\textbf{Affogato}~\cite{lee2025affogato} provides object-level 3D point clouds derived from the Objaverse ~\cite{deitke2023objaverse} assets. In \dataset, these point clouds serve as the 3D annotation surface onto which the two role-specific heatmaps are projected.
\end{itemize}

Based on these sources, we construct a large collection of 3D objects in the form
\[
(P_i, O_i, \mathcal V_i),
\]
where $P_i\subset\mathbb R^3$ is the point cloud of object $i$,
$O_i$ denotes its object category, and
$\mathcal V_i=\{v_{i,j}\}_{j=1}^{40}$ is a set of 40 views of the object.
Each view
$v_{i,j}=(I_{i,j},\Pi_{i,j},D_{i,j})$
provides a 2D observation of the object, where $I_{i,j}$ is the RGB image rendered using the camera parameters $\Pi_{i,j}$, and $D_{i,j}$ is the corresponding depth map.

As our primary goal is to construct a dataset covering diverse bimanual affordances and manipulable components, we prioritize candidate objects from three G-Objaverse categories:
\texttt{Daily-Used} (121{,}799 objects),
\texttt{Furnitures} (8{,}759 objects), and
\texttt{Electronics} (7{,}937 objects).

\subsection{Annotation Format}

For each curated object $(P_i, O_i, \mathcal V_i)$, \dataset provides a set of free-form natural-language bimanual \textit{task descriptions}
$\mathcal T_i$ and a corresponding \textit{annotation set} $\mathcal A_i$. 
Thus, each object $i$ in \dataset is represented as
\[
(P_i, O_i, \mathcal V_i, \mathcal T_i, \mathcal A_i).
\]

Specifically, let the task description set be
\[
\mathcal T_i=\{t_{i,k}\}_{k=1}^{K_i},
\]
where $K_i$ denotes its cardinality.
Each $t_{i,k}$ describes a \textit{task} generated by a VLM to describe a common manipulation in everyday human--object interaction (e.g., ``\texttt{open the bottle}'' and ``\texttt{pick up the book}'').

For embodied bimanual robot--object interaction, each $t_{i,k}$ is further translated into a cooperative \textit{role pair} $r_{i,k}$.
Following action vocabularies in human--object interaction~\cite{koppula2013affordances,chao2015semantic} and reusable skill libraries for robotic manipulation~\cite{dalal2021raps,nasiriany2022maple,jiang2024hacmanpp}, we define a hand-role vocabulary
\[
\begin{aligned}
\mathcal R^{\text{hand}} = \{&
\texttt{hold}, \texttt{lift}, \texttt{push}, \texttt{pull},
\texttt{press}, \texttt{slide}, \\
&
\texttt{rotate}, \texttt{squeeze}
\}.
\end{aligned}
\]
Each role denotes an atomic action primitive. By pairing two actions from this vocabulary, more complex everyday bimanual object manipulations can be described.
For example, the pair $(\texttt{lift}, \texttt{hold})$ can represent bimanually lifting a coffee mug, whereas $(\texttt{rotate}, \texttt{hold})$ can represent opening a bottle cap with two hands. Accordingly, we instruct a VLM to select a pair of cooperative roles
for each task $t_{i,k}$:
\[
r_{i,k}=(r_{i,k}^a,r_{i,k}^b)\in
\mathcal R^{\text{hand}} 
\times
\mathcal R^{\text{hand}} 
.
\]
The role pair is not assigned to fixed left or right hands, so that a robot can map the two actions to the effectors best suited to its embodiment.

To ground these actions on the object, we associate each role pair
$r_{i,k}$ with a \textit{heatmap pair} on $P_i$:
\[
h_{i,k}=(h_{i,k}^a,h_{i,k}^b)
.
\]
Each heatmap assigns an affordance likelihood to every point $p\in P_i$
for its corresponding role:
\[
h_{i,k}^a(p),\,h_{i,k}^b(p)\in[0,1]
.
\]

The corresponding annotation set is denoted as
\[
\mathcal A_i=
\bigl\{
(r_{i,k},h_{i,k})
\bigr\}_{k=1}^{K_i}.
\]

\subsection{Foundation Model-Based Annotation}
\label{sec:pipeline}

\cref{fig:pipeline} illustrates our annotation pipeline. Building the task--annotation pairs $(\mathcal T_i,\mathcal A_i)$ at scale for diverse objects requires
full semantic knowledge of object use and precise spatial supervision. For bimanual natural-language task
construction, we use the
open-weight multimodal model
Gemma-4-26B-A4B~\cite{gemmateam2026gemma4} to curate objects, propose tasks,
and decompose each task into two cooperative hand roles. For point-cloud spatial annotation, we employ the
pointing VLM Molmo2-8B~\cite{clark2026molmo2} to localize role-specific contact regions and the promptable segmenter SAM2~\cite{ravi2024sam2} to convert the
predicted coordinates into 2D masks. 

\paragraph{Stage 1: Object and component-usage curation.}
The objective of this stage is to select objects that are meaningful for everyday bimanual manipulation and, for each selected object, generate a set of (component, usage) pairs.
Specifically, for each object
$(P_i, O_i, \mathcal V_i)$
in the repository, we select a subset of eight semantic views
$\mathcal V_i^{\mathrm{sem}}
\subset \mathcal V_i$,
comprising one top-down view, one underside view, and six azimuth-distributed oblique views.
We use this subset instead of all 40 views to avoid overloading the model.

First, we prompt a VLM with $\mathcal V_i^{\mathrm{sem}}$ to determine if object $i$ is
suitable for everyday bimanual manipulation. Next, to ground subsequent tasks
in the object's visible structure, we instruct Gemma to
identify manipulable \textit{components} of the object and the potential \textit{uses} of these components:
\[
\mathcal C_i
=
\text{VLM}(\mathcal V_i^{\mathrm{sem}}, \text{prompt}),
\]
where $\mathcal C_i=\{x_{i,n}\}_{n=1}^{L_i}$ contains $L_i$ (component, usage) pairs. For example, the set $\mathcal C_i$ for a bottle could be
$\{(\textit{body},\textit{grasp or lift}),
(\textit{cap},\textit{turn to open}),\ldots\}$. 

We retain object $(P_i, O_i, \mathcal V_i)$ only if $|\mathcal C_i| \ge 1$ and use $\mathcal C_i$ as the textual context for task generation in
Stage~2.

\paragraph{Stage 2: Multi-granular task generation and role decomposition.}
The objective of this stage is to synthesize a set $\mathcal T_i$ of natural-language tasks from $(\mathcal C_i,\mathcal V_i^{\mathrm{sem}})$ and decompose each task $t_{i,k}\in\mathcal T_i$ into two single-hand role–query pairs,
$(r_{i,k}^a,q_{i,k}^a)$ and
$(r_{i,k}^b,q_{i,k}^b)$.

The task set $\mathcal T_i$ should cover object manipulation at multiple levels of granularity, ranging from whole-object operations~\cite{deng2021affordancenet} to component-level interactions~\cite{xu2022partafford,zhan2024oakink2} and goal-oriented functions~\cite{zhu2015understandingtools,chen2022acf}.
Specifically, \textit{object-level} tasks involve manipulating the entire object, \textit{component-level} tasks target specific object parts, and \textit{function-level} tasks specify high-level functional goals.
\cref{tab:task-families} defines these three task families and provides representative examples.
The families are distinguished by the level of abstraction of their instructions rather than by mutually exclusive physical manipulation patterns.

\begin{table}[t]
\centering
\caption{\textbf{Task families in \dataset.} Definitions and representative examples.}
\small
\setlength{\tabcolsep}{3pt}
\begin{tabular}{@{}>{\raggedright\arraybackslash}p{0.19\columnwidth}>{\raggedright\arraybackslash}p{0.375\columnwidth}>{\raggedright\arraybackslash}p{0.375\columnwidth}@{}}
\toprule
\textbf{Family} & \textbf{Definition} & \textbf{Examples} \\
\midrule
Object-level & Manipulates the entire object as a single unit. & Move a chair; rotate a table; pick up a bucket.\\
\midrule
Component-level & Manipulates a specific object component in $\mathcal C_i$. & Remove a lid; open a drawer; unscrew a cap.\\
\midrule
Function-level & Manipulates an object to achieve a high-level functional goal. & Open a book for reading; tilt a mug for drinking. \\
\bottomrule
\end{tabular}
\label{tab:task-families}
\end{table}

First, we prompt a VLM with $(\mathcal C_i,\mathcal V_i^{\mathrm{sem}})$ to generate multiple tasks across the three families, keeping only tasks that naturally require or strongly benefit from two hands (\cref{sec:annotation-prompts}). Set $\mathcal C_i$ specifies plausible object components and their usages in textual form, while the semantic views $\mathcal V_i^{\mathrm{sem}}$ provide visual context for generating object-appropriate tasks. This yields
\[
\mathcal T_i = \text{VLM}(\mathcal C_i,\mathcal V_i^{\mathrm{sem}}, \text{prompt}).
\]
Each task is then phrased as the instruction ``Point to the regions of the object you would use to $\langle t_{i,k}\rangle$.''

Next, for each $t_{i,k}\in\mathcal T_i$, we instruct a VLM with
$(t_{i,k},\mathcal C_i,\mathcal V_i^{\mathrm{sem}})$ to derive an \textit{operation plan}.
The plan contains two roles $r_{i,k}^{*}$ with $*\in\{a,b\}$, and each role $r_{i,k}^{*}$ is associated with a \textit{contact region} on a component in $\mathcal C_i$ where the hand acts, and the \textit{function} of the role in the task.
To prepare for the prompt in Stage~3, we fill a pointing query $q_{i,k}^{*}$ following the template: \texttt{Point to $\langle$\textit{contact region}$\rangle$ where a
robot hand should $\langle$\textit{role}$\rangle$ it to
$\langle$\textit{function}$\rangle$.} This yields two roles and their corresponding pointing queries:
\[
(r_{i,k},\, q_{i,k}) = \text{VLM}(t_{i,k},
\,  \mathcal C_i, \, \mathcal V_i^{\mathrm{sem}}, \text{prompt}).
\]
For example, if $t_{i,k}=$ \texttt{open the bottle}, we may have
$q_{i,k}^a=$ \texttt{Point to $\langle$the rim of the cap$\rangle$ where a robot hand should
$\langle$rotate$\rangle$ it to $\langle$actuate the cap counter-clockwise$\rangle$.} and
$q_{i,k}^b=$ \texttt{Point to $\langle$the mid-height of the body$\rangle$ where a robot hand
should $\langle$hold$\rangle$ it to $\langle$stabilize the body against rotation$\rangle$.}

\begin{figure*}[t]
\centering
\includegraphics[width=\linewidth]{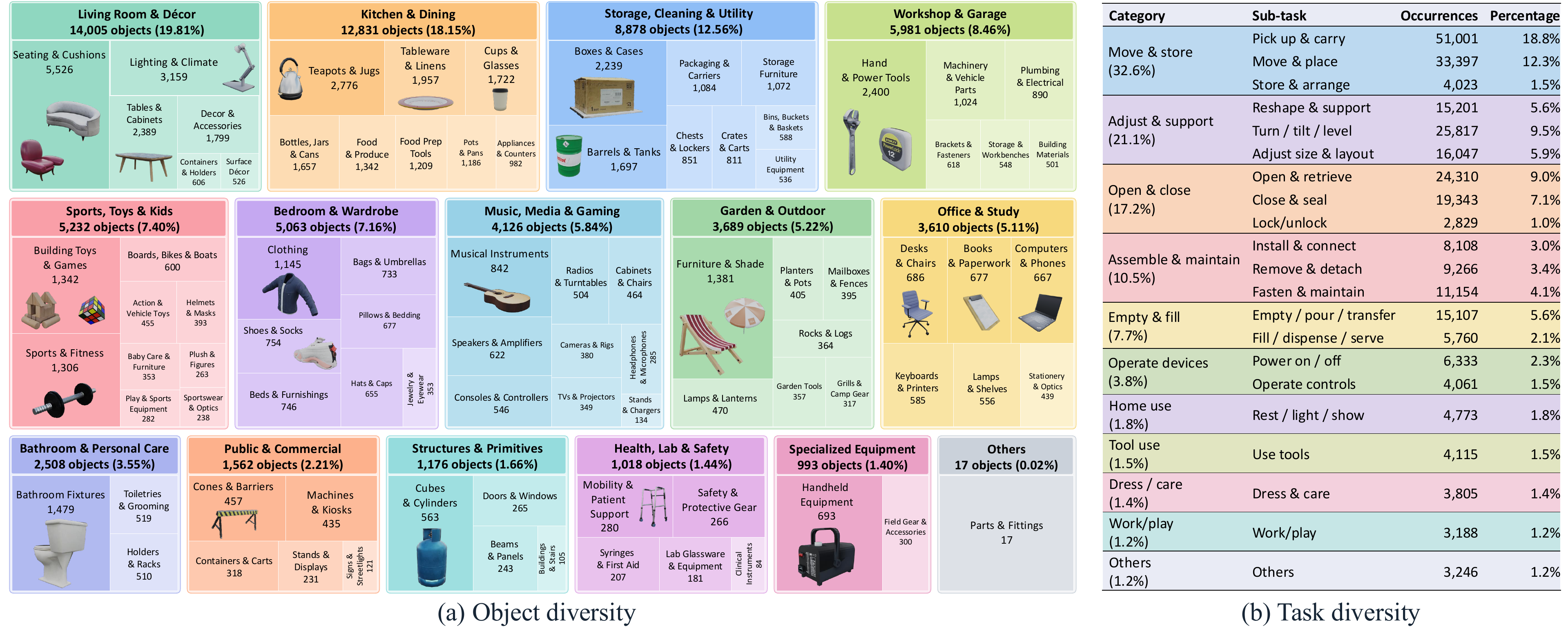}
\caption{\textbf{Object and task distributions in \dataset.}
(a) Objects by usage scenario and category; rectangle areas show category proportions per scenario, and headings give object counts and dataset shares. (b) Task counts and percentages by category and sub-task.}
\label{fig:dataset-diversity}
\vspace{-2pt}
\end{figure*}

\paragraph{Stage 3: Role-aware 2D grounding and segmentation.}
The objective of this stage is to ground the queried interaction regions in 2D images using the pointing queries constructed above.
Given a pointing query $q_{i,k}^*$ and an image $I_{i,j}$,
we use Molmo2 to localize the contact candidates:
\begin{equation}
\mathcal S_{i,k,j}^*
=
\mathrm{Molmo2}(I_{i,j}, q_{i,k}^*)
,
\label{eq:molmo2}
\end{equation}
where $\mathcal S_{i,k,j}^*$ is a variable-size set containing zero,
one, or multiple 2D point coordinates in $I_{i,j}$. This step is performed for all images $I_{i,j}$ in $\mathcal V_i$.

Although Molmo2 predicts multiple contact points for each pointing query, its outputs do not provide explicit confidence scores for individual coordinates. Prior point-to-mask pipelines, such as Affogato~\cite{lee2025affogato}, therefore retain only the first predicted coordinate. Such single-point grounding has two limitations: (i) it may produce false localizations when the queried target is \textit{occluded}, and (ii) it cannot adequately handle \textit{multiple instances} of the same target. For example, given a skateboard with four wheels or a cup with multiple straws, single-point grounding identifies only one instance, leaving the remaining instances unannotated.

To address the occlusion and multi-instance limitations, we introduce adaptive
multi-point grounding.
In \cref{eq:molmo2}, we extract all predicted contact points from Molmo2's
response, which covers multiple instances, and estimate a view-level confidence score that filters out views with an occluded target.
Specifically, Molmo2 returns the coordinates within a
\texttt{<points coords="...">} tag description, or returns \texttt{There are none}
when no target is predicted.
At the first decoding step, the logits of the tokens \texttt{<} and \texttt{There},
denoted by $\ell_{\mathrm{point}}$ and $\ell_{\mathrm{none}}$, therefore indicate the
model's preference for a pointing or a no-target response. We compute the
view-level confidence as:
\begin{equation}
c_{i,k,j}^{*}=\sigma\bigl(\ell_{\mathrm{point}}-\ell_{\mathrm{none}}\bigr),
\label{eq:existence-confidence}
\end{equation}
where $\sigma$ is the sigmoid function.
A higher $c_{i,k,j}^{*}$ indicates a stronger preference for a
pointing response over a no-target response in $I_{i,j}$, shared by all points predicted in this view.

Furthermore, the output of \cref{eq:molmo2} provides only \textit{point-level} predictions. To obtain \textit{mask-level} annotations, we employ SAM2 to generate segmentation masks:
\[
\mathcal M_{i,k,j}^{*}
=
\mathrm{SAM2}
\left(
I_{i,j},
\mathcal S_{i,k,j}^{+},
\mathcal S_{i,k,j}^{-}
\right)
,
\]
where $\mathcal S_{i,k,j}^{+}$ and $\mathcal S_{i,k,j}^{-}$ denote the
positive and negative prompts, respectively.
We invoke SAM2 separately with each positive point to obtain the set of
per-point masks
$\mathcal M_{i,k,j}^{*}
=\{M_{i,k,j}^{*,(m)}\}_{m=1}^{|\mathcal S_{i,k,j}^{*}|}$,
where each mask assigns pixel scores in $[0,1]$ and we only retain those scores above a threshold $\tau$.
As for the negative prompt, we use $\mathcal S_{i,k,j}^a$ and
$\mathcal S_{i,k,j}^b$ in a contrastive manner: when grounding
role $a$, we set
$(\mathcal S_{i,k,j}^{+},\mathcal S_{i,k,j}^{-})
=(\mathcal S_{i,k,j}^{a},\mathcal S_{i,k,j}^{b})$,
and vice versa for role $b$.
This contrastive prompting reduces cross-role leakage.
We apply this process to all 40 views in $\mathcal V_i$: Molmo2 predicts the points of a role, and SAM2 turns each point into a mask, yielding the
complete set of role-specific 2D masks.

\paragraph{Stage 4: Role-consistent 3D heatmap construction.}
The objective of this stage is to transform the above 2D masks into a heatmap pair $h_{i,k}$ on $P_i$.
First, for each role, we integrate its per-point masks
$M_{i,k,j}^{*,(m)}$ into a unified mask.
For each pixel $u$, we consolidate its score by averaging the overlaying masks:
\begin{equation}
\bar M_{i,k,j}^{*}(u)=
\frac{\sum_m M_{i,k,j}^{*,(m)}(u)}{N(u)}
,
\label{eq:mask-consolidation}
\end{equation}
where $N(u)$ is the number of masks covering pixel $u$.
The resulting $\bar M_{i,k,j}^{*}$ represents the role-specific contact
regions corresponding to the 2D view $I_{i, j}$.

Next, we project the consolidated mask $\bar M_{i,k,j}^{*}$ onto $P_i$ using the depth
$D_{i,j}$ and camera parameters $\Pi_{i,j}$. Each point $p\in P_i$ receives the
mask score $\bar M_{i,k,j}^{*}(u)$ from its corresponding pixel location
$u$ in $I_{i, j}$, where the depth map identifies the views $\mathcal V_i(p)$ in which $p$ is visible. We fuse these scores across views by
confidence-aware averaging:
\begin{equation}
h_{i,k}^{*}(p)=
\frac{1}{|\mathcal V_i(p)|}\sum_{j\in\mathcal V_i(p)} w_{i,k,j}^{*}\,\bar M_{i,k,j}^{*}(u)
,
\label{eq:confidence-weighted-voting}
\end{equation}
where $w_{i,k,j}^{*}=1$ if $c_{i,k,j}^{*}\ge 0.9$ and $0$ otherwise.

\leavevmode \cref{sec:annotation-example,sec:ablation-studies} provide an example and ablations.

\section{Dataset Statistics and Validation}
\subsection{Dataset Statistics and Diversity}
\label{sec:dataset-statistics}

\dataset contains \numObjects 3D objects and \numTasks free-form
bimanual tasks, with \numRoles role-specific surface heatmaps.
Object-level, component-level, and function-level tasks account for
\shareObjectLevelTasks, \shareComponentLevelTasks, and
\shareFunctionalPoseTasks of the annotations, respectively.

To characterize the diversity of DUALingo, \cref{fig:dataset-diversity} summarizes the distributions of objects and manipulation tasks. Objects span a wide range of usage scenarios, from household activities and sports to outdoor environments and specialized applications. Across these objects, manipulation tasks range from moving and positioning objects to adjusting, assembling, and operating them for specific purposes. Together, these distributions show broad coverage of practical bimanual interactions in diverse real-world settings.

\subsection{Qualitative Analysis of \dataset}
\label{sec:qualitative-analysis}

\leavevmode \cref{fig:teaser} visualizes representative annotations in \dataset. The contact regions follow the task and the assigned roles. For example, to pick up the cup, the \texttt{lift} role covers the handle and the \texttt{hold} role covers the body. To remove a bottle cap, the \texttt{rotate} role is confined to the small cap, while the \texttt{hold} role stays on the body. Symmetric roles occupy mirrored regions, such as the two \texttt{squeeze} roles on the bolt-cutter handles. In these examples, the annotations range from broad contacts on whole objects to fine-grained contacts on small components.

\subsection{Comparison with Existing Datasets}

\cref{tab:dataset-comparison} compares existing affordance datasets in terms of 3D representation, free-form task instructions, bimanual interaction, and role awareness. \textit{3D affordance datasets} primarily ground individual interaction regions, without modeling cooperation between two hands. In contrast, \textit{bimanual datasets} capture two-hand interactions but do not explicitly ground cooperative roles on complete 3D object surfaces. \dataset bridges this gap by providing free-form, task-conditioned annotations of two cooperative roles, each grounded as a heatmap on the 3D object surface.

\subsection{Annotation Quality and Data Splits}
\label{sec:annotation-quality}

Exhaustive human review is impractical at the scale of \dataset. Following the LLM-as-a-judge protocols~\cite{liu2023geval,zheng2023judging}, we employ the vision-language model Qwen3.8-27B~\cite{qwen38} as a multimodal judge to systematically evaluate annotation quality using predefined rating criteria.

\noindent\textbf{Judge validation.} To examine whether the judge aligns with human judgment, we randomly sample
\numHumanTasks object-task annotations and ask
five experts to independently rate \emph{task validity},
\emph{cooperative role validity}, and \emph{paired heatmap quality}.
Ratings range from bad (0) to okay (1, minor flaws) and good (2).
The judge receives object views, task and role descriptions, and paired heatmaps, and assigns ratings based on consistent criteria.
As shown in \cref{tab:human-quality}, the judge's mean
quality scores closely match expert ratings on the sampled annotations,
and it agrees with individual experts as closely as the experts agree with one another,
making it a reliable and scalable proxy for expert assessment.

We apply the same judge to the full corpus, obtaining scores of 92.5, 95.7, and 89.7 for task, role, and heatmap quality, respectively. These differ from the sampled evaluation by 0.8, 0.8, and 0.6 points, indicating consistent annotation quality at scale. Detailed protocols are provided in \cref{sec:quality-protocol}.

\begin{table}[t]
\centering
\caption{\textbf{Annotation quality assessment.} Human (1K) averages five expert ratings; LLM (1K) evaluates the same samples; LLM (All) evaluates the full corpus. Scores are normalized to 0--100. AC1~\cite{gwet2008ac1}: judge--expert (J--E) and expert--expert (E--E) agreement.}
\label{tab:human-quality}
\small
\setlength{\tabcolsep}{2pt}
\begin{tabular}{@{}lccccc@{}}
\toprule
& \multicolumn{3}{c}{Score} & \multicolumn{2}{c}{AC1} \\
\cmidrule(l{2pt}r{2pt}){2-4}\cmidrule(l{2pt}r{0pt}){5-6}
Criterion & Human (1K) & LLM (1K) & LLM (All) & J--E & E--E \\
\midrule
Task validity & 94.1 & 93.3 & 92.5 & 0.82 & 0.81 \\
Role validity & 97.6 & 96.5 & 95.7 & 0.93 & 0.93 \\
Heatmap quality & 87.4 & 90.3 & 89.7 & 0.74 & 0.74 \\
\bottomrule
\end{tabular}
\end{table}

\noindent\textbf{Release and data splits.}
We exclude annotations rated bad by the judge on any criterion (\shareRemovedBad of the generated annotations). Among the retained sampled annotations, most experts rate 0.1\%, 0\%, and 2.0\% as bad on the three criteria. We split the remaining data by object into training (94\%), validation (1\%), and test (5\%) sets, stratified by source category, with no object shared across splits.

\begin{figure*}[t]
\centering
\includegraphics[width=\textwidth]{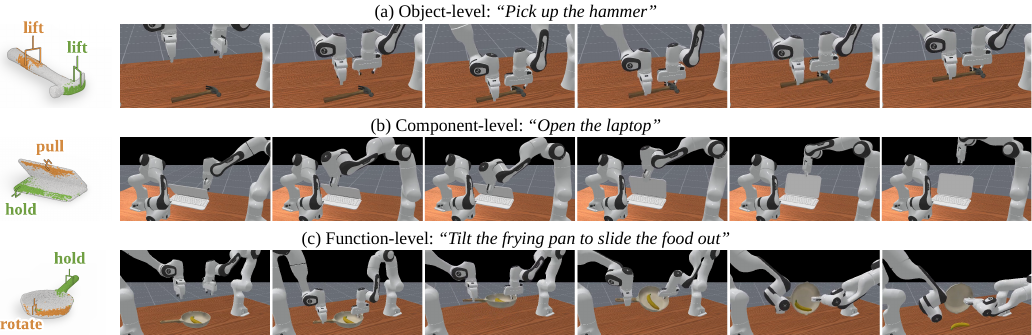}
\caption{\textbf{Bimanual manipulation on unseen objects.} Dual-Espresso-3D, trained on \dataset, is deployed in ManiSkill3 on objects from YCB, OakInk2, and PhysX-Mobility. Left: top 10\% highest-scoring heatmap regions. Right: manipulation sequences.}
\label{fig:sim-rollouts}
\end{figure*}

\section{Benchmarking and Robotic Demonstration}
\label{sec:experiments}

\subsection{Implementation and Baselines}

During dataset curation, we accelerate Gemma and Molmo2 inference using vLLM~\cite{kwon2023efficient} with BF16 precision and greedy decoding. We set the mask threshold to $\tau=0.1$. The pipeline runs on two NVIDIA RTX PRO 6000 Blackwell Max-Q GPUs, consuming 1{,}060 GPU-hours in total, i.e., under one GPU-minute per object.

We establish baselines from three representative 3D affordance grounding models that take free-form instructions and predict point-wise heatmaps, OpenAD~\cite{nguyen2023openad}, PointRefer~\cite{li2024laso}, and Espresso-3D~\cite{lee2025affogato}, in two settings.
\begin{itemize}
\item 
\emph{Zero-shot evaluation:}
We assess whether pretrained single-region affordance models transfer to bimanual tasks. For each task, we query each model once per role. Each query explicitly specifies both roles:
``Point to the part you would $\langle r^a\rangle$ while your other hand $\langle r^b\rangle$, to $\langle t\rangle$.''
\item
\emph{Training on \dataset:}
We adapt the three single-region models for bimanual affordance grounding, denoting them \textbf{Dual-OpenAD}, \textbf{Dual-PointRefer}, and \textbf{Dual-Espresso-3D}. Each adaptation adds two learnable role embeddings and a lightweight role head while preserving the original architecture. Given only the task instruction (\cref{sec:pipeline}), the adapted model predicts the cooperative role pair and two role-specific heatmaps. We fine-tune the pretrained models on \dataset and report results averaged over three random seeds.
\end{itemize}
\cref{sec:benchmark-details} provides details on the adapted architectures, training, and evaluation.

\subsection{Benchmarking on \dataset}
\label{sec:benchmark}

\begin{table}[t]
\centering
\caption{\textbf{Benchmark results on DUALingo.} Original models are evaluated zero-shot, while adapted Dual models (shaded) are fine-tuned on DUALingo. Best results are in bold.}
\label{tab:benchmark}
\small
\setlength{\tabcolsep}{2pt}
\begin{tabularx}{\linewidth}{l*{4}{>{\centering\arraybackslash}X}c}
\toprule
Method & aIoU\,$\uparrow$ & AUC\,$\uparrow$ & SIM\,$\uparrow$ & MAE\,$\downarrow$ & Role\,$\uparrow$ \\
\midrule
OpenAD~\cite{nguyen2023openad}     & 3.9  & 50.4 & 37.9 & 0.332 & -- \\
\rowcolor{gray!12}
Dual-OpenAD                        & 25.4 & 70.4 & 53.7 & 0.275 & 73.1 \\
\midrule
PointRefer~\cite{li2024laso}       & 7.3  & 51.3 & 32.7 & 0.332 & -- \\
\rowcolor{gray!12}
Dual-PointRefer                    & 34.8 & 78.6 & 61.8 & 0.198 & \textbf{79.1} \\
\midrule
Espresso-3D~\cite{lee2025affogato} & 10.3 & 53.7 & 39.8 & 0.331 & -- \\
\rowcolor{gray!12}
Dual-Espresso-3D                   & \textbf{37.8} & \textbf{79.5} & \textbf{62.8} & \textbf{0.191} & 78.8 \\
\bottomrule
\end{tabularx}
\end{table}

Following LASO~\cite{li2024laso} and Affogato~\cite{lee2025affogato}, we report
average Intersection over Union (aIoU), Area Under the ROC Curve (AUC),
Similarity (SIM), and Mean Absolute Error (MAE) between the predicted and
annotated heatmaps. Heatmaps are compared role by role. When the two roles are
identical, we take the better of the two assignments.
Additionally, we report role
accuracy (Role), the percentage of tasks for which both roles are
predicted correctly.

\cref{tab:benchmark} reports the results on the test set. In the zero-shot evaluation, all three models perform poorly despite explicit specification of both roles, with AUC scores of 50.4--53.7, only slightly above the random-guessing baseline of 50. Moreover, the two predicted role heatmaps largely overlap, with an average IoU around 0.6 between their top 10\% highest-scoring points (\cref{sec:benchmark-results}), indicating that the models struggle to distinguish between the two roles. These results show that single-region affordance knowledge transfers poorly to bimanual grounding even when both roles are given, highlighting the need for role-aware supervision.

Training on \dataset improves all metrics across the three adapted models, while producing nearly disjoint role heatmaps (\cref{sec:benchmark-results}). Dual-Espresso-3D achieves the best grounding performance, increasing aIoU from 10.3 to 37.8 and correctly predicting both roles for 78.8\% of test tasks. Consistent gains across three architectures, which also hold on expert-verified test tasks (\cref{sec:benchmark-results}), demonstrate the value of \dataset for learning bimanual affordances through adaptations of existing single-region models. Nevertheless, role accuracy in \cref{tab:benchmark} remains below 80\% and AUC below 80 for all models, indicating substantial room for improvement beyond these adaptations.

\subsection{Bimanual Manipulation on Unseen Objects}
\label{sec:simulation}

We illustrate how models trained on \dataset{} transfer to unseen objects and support robotic manipulation by deploying Dual-Espresso-3D in ManiSkill3~\cite{tao2025maniskill3}. We select objects from three external sources: YCB~\cite{calli2015ycb}, OakInk2~\cite{zhan2024oakink2}, and PhysX-Mobility~\cite{cao2026physxanything}.
Dual-Espresso-3D predicts task-appropriate contact regions on these unseen objects (\cref{fig:sim-rollouts}). To lift a hammer, the two \texttt{lift} roles target its handle and head, respectively. For an articulated laptop, the top 10\% highest-scoring points of the \texttt{pull} and \texttt{hold} heatmaps lie on the screen and base, respectively. To tilt a frying pan, the \texttt{rotate} role targets the rim while the \texttt{hold} role targets the handle. More examples are in \cref{sec:unseen-results}.

To assess whether the predicted heatmaps are actionable, we use them to guide two Franka robot arms with scripted controllers. In physics simulation, the arms successfully complete all three tasks: lifting the hammer approximately 15 cm while keeping it level, opening the laptop screen from $45^\circ$ to $110^\circ$ while stabilizing the base, and tilting the pan by $96^\circ$ to slide its contents onto the table. These demonstrations show that the predicted contact regions can guide coordinated bimanual manipulation on unseen objects.

\section{Conclusion}
\label{sec:conclusion}

We introduced \dataset, a large-scale dataset for cooperative bimanual affordance grounding on everyday 3D objects. Using only open-weight foundation models, we generate multi-granular tasks, decompose them into hand-agnostic cooperative roles, and ground each role as a 3D surface heatmap. Across diverse objects and tasks, \dataset provides a foundation for training models that reason about where two hands make contact and in which roles. We envision such models transferring cooperative skills across robot embodiments toward general-purpose robotic assistance in everyday environments.

\section{LLM Usage Statement}

\noindent\textbf{Use in the research.}
Generative AI models are a core component of the dataset construction in \cref{sec:pipeline}. The open-weight multimodal model Gemma-4-26B-A4B~\cite{gemmateam2026gemma4} curates objects, identifies their manipulable components, generates the natural-language tasks, and decomposes each task into a role pair and pointing queries (Stages~1--2). The pointing model Molmo2-8B~\cite{clark2026molmo2} localizes contact points in the rendered views, and its first-step token logits provide the view-level confidence (Stage~3). The promptable segmentation model SAM2~\cite{ravi2024sam2} converts these points into masks. Qwen3.8-27B~\cite{qwen38} serves as the multimodal judge that assesses annotation quality and filters the released data (\cref{sec:annotation-quality}). All models are used with their released weights and without fine-tuning, and their outputs determine the released tasks, roles, and heatmaps. Human experts independently rate a random sample of \numHumanTasks annotations to validate these outputs and the judge. No LLM produces the expert ratings or the benchmark results.

\noindent\textbf{Use in writing and code.}
We used Claude (Anthropic), through Claude Code, to edit the grammar and wording of the paper and to help draft text from the authors' outlines and results. We also used it as a coding assistant to implement the annotation pipeline, the judge, the benchmark, and the simulation, to refine the model prompts, and to plot figures. We also used Claude and GPT (OpenAI) to obtain reviewer-style feedback on drafts. The research design and conclusions are the authors' own. The authors reviewed and verified all AI-assisted text, code, and citations, and take full responsibility for the content of this paper.

{
    \small
    \bibliographystyle{ieeenat_fullname}
    \bibliography{main}
}

\clearpage
\twocolumn[{%
\centering
{\Large\textbf{\thetitle}\\
\vspace{0.5em}Supplementary Material\\
\vspace{1.0em}}
}]
\appendix
\crefalias{section}{appendix}\crefalias{subsection}{subappendix}
\crefname{appendix}{Appendix}{Appendices}\crefname{subappendix}{Appendix}{Appendices}
\section{Annotation Example}
\label{sec:annotation-example}

\begin{figure*}[t]
\centering
\includegraphics[width=\linewidth]{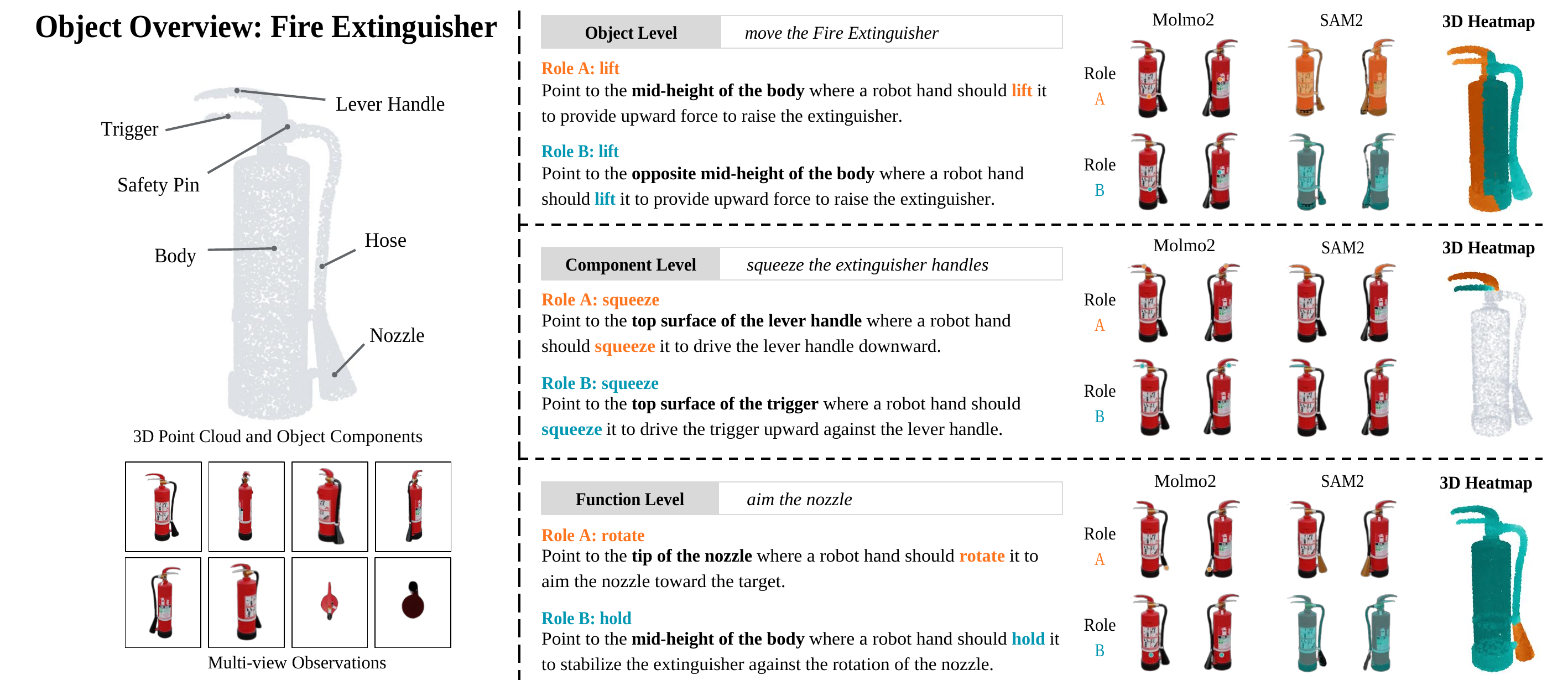}
\caption{\textbf{Annotation of a fire extinguisher.} Left: the input point cloud and views, with the manipulable components found in Stage~1. Right: for one task of each family, the role pair and pointing queries (Stage~2), the points and masks in two views (Stage~3), and the heatmap pair (Stage~4). Role $a$ is orange and role $b$ is teal.}
\label{fig:annotation_examples}
\end{figure*}

\cref{fig:annotation_examples} follows one object, a fire extinguisher, through
the four stages of \cref{sec:pipeline}. From eight views, Stage~1
identifies its manipulable components, including the lever handle, the trigger,
the safety pin, and the nozzle (left). For one task of each family, Stage~2
derives the role pair and a pointing query per role, which names the contact
region and the purpose of the contact (right). Stage~3 grounds each query as
points with Molmo2 and as masks with SAM2, shown here in two of the 40 views.
Stage~4 fuses these masks into the heatmap pair on the point cloud.

The three tasks show how the role pair and the contacts change with the task on
a single object. To move the extinguisher, two \texttt{lift} roles hold opposite
sides of the body. To squeeze its handles, two \texttt{squeeze} roles press the
lever down and the trigger up. To aim the nozzle, a \texttt{rotate} role turns
the nozzle tip while a \texttt{hold} role keeps the body still. \cref{sec:annotation-prompts} shows the prompts of Stages~1 and~2.

\paragraph{Object and task counts.}
Of the 138{,}495 candidate objects, Stage~1 retains 71{,}389 (51.5\%), for which
Stage~2 generates 311{,}714 tasks. Stages~3 and~4 produce heatmaps for 310{,}234 of
them, the judge removes 12.6\% (\cref{sec:annotation-quality}), and 276 tasks with a
role outside the vocabulary are discarded. \dataset then contains \numTasksExact tasks on
\numObjectsExact objects, an average of 3.8 tasks per object.

\section{Ablations of the Annotation Pipeline}
\label{sec:ablation-studies}

\paragraph{Adaptive multi-point grounding.}
To assess the effect of adaptive multi-point grounding on target coverage,
we compare first-point and multi-point selection using the same Molmo2
responses. The baseline selects one coordinate per view, whereas adaptive
grounding keeps every coordinate that Molmo2 predicts, so the number of points adapts to each view.

\begin{figure}[!htb]
\centering
\includegraphics[width=\linewidth]{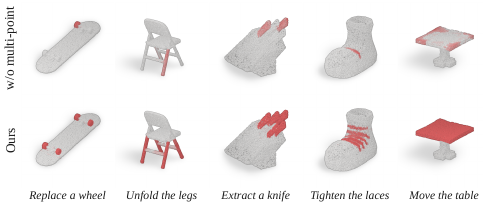}
\caption{\textbf{Effect of adaptive multi-point grounding.}
Final 3D heatmaps for five tasks, without (top) and with (bottom)
multi-point grounding. Red highlights the annotated contact regions.}
\label{fig:adaptive-grounding-ablation}
\end{figure}

\cref{fig:adaptive-grounding-ablation} shows incomplete coverage
under single-point selection, even after multi-view aggregation.
For example, the baseline annotates only one skateboard wheel and misses
several chair legs and lace segments.
In contrast, adaptive multi-point grounding covers all four wheels and
recovers the omitted leg and lace regions.
Beyond repeated targets, the tabletop comparison shows more continuous
coverage of an extended contact region.
Together, the examples demonstrate the importance of grounding multiple
contact candidates for more complete 3D annotations.

\paragraph{Role-aware segmentation.}
To assess the contribution of partner-role negative prompts, we compare segmentation with and without these prompts.

\begin{figure}[!htb]
\centering
\includegraphics[width=\linewidth]{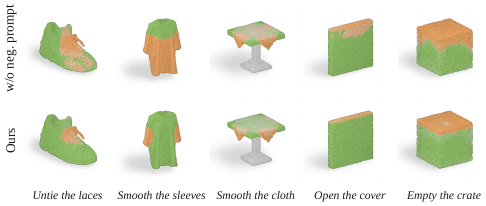}
\caption{\textbf{Effect of partner-role negative prompts.}
Final paired 3D heatmaps without (top) and with (bottom) negative prompts.
Orange and green distinguish the two cooperative roles.}
\label{fig:role-aware-segmentation}
\end{figure}

\cref{fig:role-aware-segmentation} shows the effect on the final
paired 3D heatmaps.
Without negative prompts, the shoelace annotation extends over the shoe
body, and the sleeve annotation spreads onto the garment's main body.
With partner-role negatives, the annotations follow the shoelaces and
sleeves more closely and separate them from the partner-role regions.
The comparison demonstrates that positive points alone can be insufficient
to distinguish adjacent interaction regions; partner-role negatives reduce
cross-role leakage and preserve more distinct contact regions in the final
3D annotations.

\section{Annotation Quality Protocol}
\label{sec:quality-protocol}

\begin{figure*}[t]
\centering
\includegraphics[width=\textwidth]{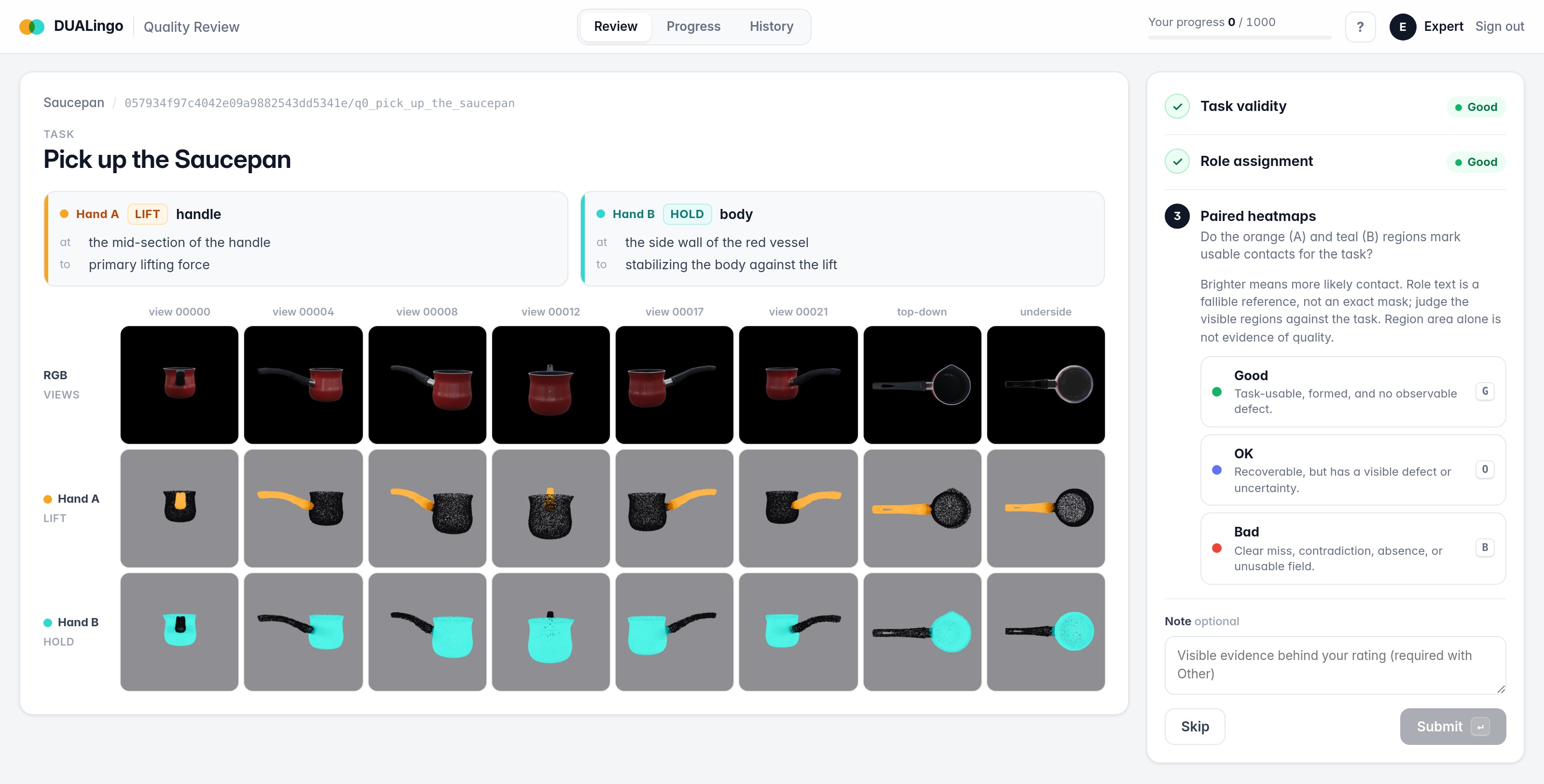}
\caption{\textbf{Expert rating interface.} The expert sees eight views of the
object, the task, the two roles, and the paired heatmaps, and rates task validity,
role validity, and heatmap quality in turn.}
\label{fig:review-ui}
\end{figure*}

\paragraph{Expert rating.}
We recruit five human experts, all graduate students in computer vision and
robotics, to rate the \numHumanTasks sampled annotations in a web interface
(\cref{fig:review-ui}). The experts and the judge rate the same three
criteria, task validity, role validity, and heatmap quality, on the same
three-point scale of bad (0), okay (1), and good (2). The middle rating marks an
annotation that is neither clearly good nor clearly bad. The task criterion asks whether the task is a
sensible two-handed task for the object. Each expert rates every annotation
independently, without access to the ratings of the other experts or the
outputs of the judge. The interface reveals the evidence step by step: the task
is rated before the roles are shown, and the roles before the heatmaps, so that
each criterion is judged on its own. A rating of bad or okay requires at least
one reason tag.

\paragraph{Judge.}
The judge is Qwen3.8-27B with one fixed prompt per criterion
(\cref{fig:judge-prompt-task,fig:judge-prompt-role,fig:judge-prompt-heatmap}),
which covers the same three criteria on the same scale, and it is
decoded greedily into a fixed JSON format.

\paragraph{Agreement with experts.}
We measure agreement with Gwet's AC1~\cite{gwet2008ac1}, which, unlike Cohen's
kappa, remains stable when most annotations are rated good.
\cref{tab:judge-agreement} shows that the judge closely matches the experts' mean
scores and agrees with individual experts as closely as the experts agree with
one another. The judge is therefore a reliable proxy for expert assessment at
scale.

\paragraph{Released annotations.}
The scores in \cref{tab:human-quality} describe all generated annotations, before
any removal, whereas the statistics in \cref{sec:dataset-statistics} describe the
released data. Between the two, the judge acts as an automatic supervisor over
the full corpus and screens out the annotations it rates bad.
\cref{tab:expert-distribution} reports how the experts rate the 883 sampled
annotations that remain, where each annotation takes the median of its five
ratings. Of these, 98.8\%, 99.8\%, and 88.4\% are good for task validity, role
validity, and heatmap quality, and 0.1\%, 0\%, and 2.0\% are bad. These results
confirm the high annotation quality of \dataset across all splits, and
\cref{sec:benchmark-results} further reports consistent results on
expert-verified test tasks.

\begin{table}[!htb]
\centering
\caption{\textbf{Agreement between the judge and the experts} on the
\numHumanTasks sampled annotations. Scores rescale the mean rating to
0--100, with the standard deviation across experts. AC1 is averaged over judge--expert (J--E) and
expert--expert (E--E) pairs.}
\label{tab:judge-agreement}
\small
\setlength{\tabcolsep}{5pt}
\begin{tabular}{@{}lcccc@{}}
\toprule
& \multicolumn{2}{c}{Score} & \multicolumn{2}{c}{AC1} \\
\cmidrule(l{5pt}r{5pt}){2-3}\cmidrule(l{5pt}r{0pt}){4-5}
Criterion & Experts & Judge & J--E & E--E \\
\midrule
Task validity & 94.1\,{\scriptsize$\pm$3.5} & 93.3 & 0.82 & 0.81 \\
Role validity & 97.6\,{\scriptsize$\pm$1.1} & 96.5 & 0.93 & 0.93 \\
Heatmap quality & 87.4\,{\scriptsize$\pm$3.4} & 90.3 & 0.74 & 0.74 \\
\bottomrule
\end{tabular}
\end{table}

\begin{table}[!htb]
\centering
\caption{\textbf{Expert ratings of the retained annotations.} Share of the 883 sampled annotations retained by the judge that are good, okay, or bad, where each annotation takes the median of its five expert ratings.}
\label{tab:expert-distribution}
\small
\begin{tabular}{@{}lccc@{}}
\toprule
Criterion & Good & Okay & Bad \\
\midrule
Task validity & 98.8\% & 1.1\% & 0.1\% \\
Role validity & 99.8\% & 0.2\% & 0.0\% \\
Heatmap quality & 88.4\% & 9.5\% & 2.0\% \\
\bottomrule
\end{tabular}
\end{table}

\section{Benchmark Details}
\label{sec:benchmark-details}

This section describes how we adapt prior single-region models to \dataset
(\cref{sec:benchmark-models}), how we train (\cref{sec:benchmark-training}) and
evaluate them (\cref{sec:benchmark-eval}), and additional results
(\cref{sec:benchmark-results}). For brevity, we drop the indices $i,k$: a task
$t$ has the role pair $(r^a,r^b)$ and the heatmap pair $(h^a,h^b)$.

\subsection{Adapting Prior Models}
\label{sec:benchmark-models}

All three models predict a single heatmap for a text instruction. We adapt
them to the two roles of \dataset with a shared design that changes only their
query and adds a role head. The embedding $\mathbf t$ of the task instruction
yields two role-specific queries, $\mathbf t+\mathbf e_a$ and
$\mathbf t+\mathbf e_b$, where $\mathbf e_a$ and $\mathbf e_b$ are learnable
role embeddings. Each query predicts the heatmap of its role with the original
prediction head, and a lightweight role head shared by the two queries predicts
its role from $\mathcal R^{\text{hand}}$. The role embeddings let the two
queries, which share the same task, predict different heatmaps. The Dual models
receive only the task instruction ``Point to the regions of the object you
would use to $\langle t\rangle$.'' \cref{tab:dual-models} summarizes the three
adapted models.

\begin{table*}[t]
\centering
\caption{\textbf{Adapting prior models.} The task embedding $\mathbf t$ plus a
learnable role embedding $\mathbf e_*$ forms the query of role $*\in\{a,b\}$.
All other modules and their pretrained weights are preserved.}
\label{tab:dual-models}
\footnotesize
\setlength{\tabcolsep}{4pt}
\renewcommand{\arraystretch}{1.15}
\begin{tabular}{@{}p{0.14\textwidth}p{0.255\textwidth}p{0.255\textwidth}p{0.255\textwidth}@{}}
\toprule
 & Dual-OpenAD & Dual-PointRefer & Dual-Espresso-3D \\
\midrule
Task embedding $\mathbf t$ & CLIP embedding of the instruction & mean of the token embeddings & projected sentence embedding \\
Role-specific queries & labels of the point classifier, with \texttt{none} & decoder queries, appended to the token queries & decoder queries, replacing the sentence query \\
Role head & MLP on $\mathbf t+\mathbf e_*$ & linear layer on the decoder output & linear layer on the decoder output \\
\bottomrule
\end{tabular}
\end{table*}

The three models differ only in where the role-specific queries enter.
OpenAD~\cite{nguyen2023openad} classifies each point against the CLIP embeddings
of a label set, which we replace with the two role-specific queries and a
\texttt{none} label, so the heatmap of each role is the probability of its label.
PointRefer~\cite{li2024laso} decodes the token embeddings of the question over
point features. We take their mean as $\mathbf t$ and append the two
role-specific queries to the decoder queries.
Espresso-3D~\cite{lee2025affogato} decodes a sentence query over
PartField~\cite{liu2025partfield} features. We replace the sentence query with
the two role-specific queries, which attend to each other through the
self-attention of the decoder.

\subsection{Training}
\label{sec:benchmark-training}

\paragraph{Data and hyperparameters.}
We train on the training set (\numTrainTasks tasks on \numTrainObjects objects)
for 20 epochs, with batches of 16 objects and five tasks per object. The
splits are stratified by the three G-Objaverse categories of
\cref{sec:source-assets-main} and share no object, so the test set measures
generalization to unseen objects. We use
AdamW~\cite{loshchilov2019adamw} with a learning rate of $10^{-4}$
($5\times10^{-6}$ for the fine-tuned text encoders), a weight decay of
$10^{-3}$, a linear warm-up over 300 iterations, and a cosine decay. We
evaluate on the validation set every half epoch and test the checkpoint with
the best validation aIoU. All weights are updated, except the CLIP text encoder of Dual-OpenAD and the PartField encoder of Dual-Espresso-3D, which stay frozen as in the original models. Across the three seeds, the standard deviation in \cref{tab:benchmark} is below 0.3 for aIoU, 0.3 for AUC, 0.2 for SIM, 0.002 for MAE, and 0.3 for role accuracy.

\paragraph{Losses.}
Dual-PointRefer and Dual-Espresso-3D apply the binary cross-entropy and Dice
losses of LASO~\cite{li2024laso} to the heatmap of each role, and Dual-OpenAD
keeps the class-weighted cross-entropy loss of OpenAD over the labels
$\{a,b,\texttt{none}\}$. A cross-entropy loss on the two roles is added with a
weight of 0.5. When the two roles are identical, their contacts are
interchangeable, and the loss takes the better of the two assignments of
predictions to annotations.

\subsection{Evaluation}
\label{sec:benchmark-eval}

\paragraph{Zero-shot instructions.}
Each original model is queried once per role. For a task with distinct roles,
the instruction names the queried role and its partner, e.g., ``Point to the
part you would pull while your other hand holds, to unfold the cloth,'' and the
instruction for the other role swaps the two. For a task with identical roles
$r$, we use a single instruction, ``Point to the parts you would
$\langle r\rangle$ with both hands to $\langle t\rangle$.'' OpenAD receives the
label set $\{\text{instruction},\texttt{none}\}$ and predicts the probability of
the instruction label, whereas PointRefer and Espresso-3D take the instruction
as their text input. The instructions name the roles but never the contact
regions. In contrast, the Dual models receive only the task and predict the
roles themselves.

\paragraph{Heatmap metrics.}
Each model outputs one heatmap per role, and each predicted heatmap is compared
with the annotated heatmap of the same role. When the two roles are identical,
the two predicted heatmaps are compared with the two annotated heatmaps under
the better of the two assignments, i.e., the one with the lower training loss
(\cref{sec:benchmark-training}). A zero-shot model predicts a single heatmap
for such a task, which is compared with each annotated heatmap. Following
LASO~\cite{li2024laso} and Affogato~\cite{lee2025affogato}, aIoU averages the IoU
over 20 thresholds evenly spaced in $[0,1]$, with each annotated heatmap
binarized at 0.5. AUC is the area under the ROC curve, SIM is the histogram
intersection between the normalized heatmaps, and MAE is the mean absolute error
over points. All metrics are computed on the \numPoints points of each point
cloud, and the predictions of OpenAD and PointRefer, which operate on 2{,}048
points, are interpolated to all points.

\paragraph{Role accuracy.}
Role accuracy is the percentage of tasks for which both predicted roles match
the annotated pair. The pair is ordered as annotated, with the acting role
before a stabilizing one, e.g., $(\texttt{rotate},\texttt{hold})$, and the two
outputs of a Dual model follow the same order. Role labels and heatmaps are
scored independently: each predicted heatmap is compared with the annotated
heatmap of the corresponding role, whatever role label the model predicts.

\subsection{Additional Results}
\label{sec:benchmark-results}

\paragraph{Expert-verified test tasks.}
Following Affogato~\cite{lee2025affogato}, we also evaluate models on
expert-verified test tasks. The test split includes the objects of the \numHumanTasks annotations rated by the experts
(\cref{sec:annotation-quality}).
Among the 883 that are retained, the 776 test tasks that most experts rate good on all three criteria
form an expert-verified subset. \cref{tab:benchmark-verified} compares the results
on this subset with those on the full test split: the rankings and the gains from
training are consistent.

\paragraph{Role pairs.}
The 14{,}059 test tasks span 27 role pairs, led by $(\texttt{rotate}, \texttt{hold})$
at 29.0\%, $(\texttt{lift}, \texttt{lift})$ at 22.1\%, and $(\texttt{pull}, \texttt{hold})$
at 13.4\%. Always predicting the most frequent pair yields 29.0\% role accuracy,
whereas every Dual model exceeds 73\% after fine-tuning (\cref{tab:benchmark}).

\paragraph{Distinct and identical roles.}
\cref{tab:benchmark-roletype} reports the results separately for the 9{,}502 test
tasks with distinct roles and the 4{,}557 test tasks with identical roles. The
gains in \cref{tab:benchmark} hold on both sets, including the distinct-role
tasks, which involve no assignment. On distinct-role tasks, the Dual models
raise aIoU from 4.3, 8.0, and 11.5 to 26.6, 36.1, and 39.7. On identical-role
tasks, they raise it from 3.1, 5.9, and 7.9 to 22.9, 32.1, and 33.9. Role
accuracy on each set stays within one point of that on the full test split.

\begin{table*}[t]
\centering
\caption{\textbf{Results by role type.} The test split has 9{,}502 tasks with
distinct roles and 4{,}557 tasks with identical roles. Models and training are as
in \cref{tab:benchmark}. Best results in each set are in bold.}
\label{tab:benchmark-roletype}
\small
\setlength{\tabcolsep}{5pt}
\begin{tabular}{lcccccccccc} %
\toprule
& \multicolumn{5}{c}{Distinct roles (9{,}502 tasks)} & \multicolumn{5}{c}{Identical roles (4{,}557 tasks)} \\
\cmidrule(lr){2-6}\cmidrule(lr){7-11}
Method & aIoU\,$\uparrow$ & AUC\,$\uparrow$ & SIM\,$\uparrow$ & MAE\,$\downarrow$ & Role\,$\uparrow$ & aIoU\,$\uparrow$ & AUC\,$\uparrow$ & SIM\,$\uparrow$ & MAE\,$\downarrow$ & Role\,$\uparrow$ \\
\midrule
OpenAD~\cite{nguyen2023openad}     & 4.3  & 50.4 & 38.4 & 0.352 & --   & 3.1  & 50.3 & 37.0 & 0.290 & -- \\
\rowcolor{gray!12}
Dual-OpenAD                        & 26.6 & 67.4 & 53.5 & 0.273 & 73.3 & 22.9 & 76.7 & 54.2 & 0.280 & 72.8 \\
\midrule
PointRefer~\cite{li2024laso}       & 8.0  & 51.6 & 35.9 & 0.349 & --   & 5.9  & 50.8 & 26.1 & 0.297 & -- \\
\rowcolor{gray!12}
Dual-PointRefer                    & 36.1 & 77.6 & 60.9 & 0.192 & \textbf{79.5} & 32.1 & \textbf{80.6} & 63.8 & 0.211 & 78.2 \\
\midrule
Espresso-3D~\cite{lee2025affogato} & 11.5 & 53.8 & 40.4 & 0.349 & --   & 7.9  & 53.6 & 38.7 & 0.295 & -- \\
\rowcolor{gray!12}
Dual-Espresso-3D                   & \textbf{39.7} & \textbf{79.0} & \textbf{61.9} & \textbf{0.185} & 78.7 & \textbf{33.9} & \textbf{80.6} & \textbf{64.6} & \textbf{0.203} & \textbf{78.9} \\
\bottomrule
\end{tabular}
\end{table*}

\begin{table*}[t]
\centering
\caption{\textbf{Results on the full test split and the expert-verified subset.}
The full test split has 14{,}059 tasks; the expert-verified subset has the 776 test
tasks that most experts rate good on all three criteria. The Dual models (shaded)
are fine-tuned as in \cref{tab:benchmark}. Best results in each set are in bold.}
\label{tab:benchmark-verified}
\small
\setlength{\tabcolsep}{5pt}
\begin{tabular}{lcccccccccc} %
\toprule
& \multicolumn{5}{c}{Full test (14{,}059 tasks)} & \multicolumn{5}{c}{Expert-verified (776 tasks)} \\
\cmidrule(lr){2-6}\cmidrule(lr){7-11}
Method & aIoU\,$\uparrow$ & AUC\,$\uparrow$ & SIM\,$\uparrow$ & MAE\,$\downarrow$ & Role\,$\uparrow$ & aIoU\,$\uparrow$ & AUC\,$\uparrow$ & SIM\,$\uparrow$ & MAE\,$\downarrow$ & Role\,$\uparrow$ \\
\midrule
OpenAD~\cite{nguyen2023openad}     & 3.9  & 50.4 & 37.9 & 0.332 & --   & 4.0  & 50.2 & 38.5 & 0.337 & -- \\
\rowcolor{gray!12}
Dual-OpenAD                        & 25.4 & 70.4 & 53.7 & 0.275 & 73.1 & 26.2 & 72.3 & 54.0 & 0.284 & 77.4 \\
\midrule
PointRefer~\cite{li2024laso}       & 7.3  & 51.3 & 32.7 & 0.332 & --   & 7.5  & 50.9 & 31.7 & 0.338 & -- \\
\rowcolor{gray!12}
Dual-PointRefer                    & 34.8 & 78.6 & 61.8 & 0.198 & \textbf{79.1} & 36.6 & 80.1 & 63.7 & 0.203 & \textbf{83.7} \\
\midrule
Espresso-3D~\cite{lee2025affogato} & 10.3 & 53.7 & 39.8 & 0.331 & --   & 11.2 & 53.3 & 40.7 & 0.337 & -- \\
\rowcolor{gray!12}
Dual-Espresso-3D                   & \textbf{37.8} & \textbf{79.5} & \textbf{62.8} & \textbf{0.191} & 78.8 & \textbf{39.4} & \textbf{80.7} & \textbf{64.7} & \textbf{0.196} & 83.5 \\
\bottomrule
\end{tabular}
\end{table*}

\paragraph{Training from scratch.}
\cref{tab:benchmark-scratch} compares fine-tuning from the pretrained weights
with training from scratch, which keeps only generic pretrained encoders. The
two perform similarly, within 1.7 aIoU, and both exceed the zero-shot models by
at least 21 aIoU. The gains in \cref{tab:benchmark} therefore come from the
supervision in \dataset rather than from single-region pretraining.

\begin{table}[t]
\centering
\caption{\textbf{Training from scratch vs.\ fine-tuning.} The Dual models
(shaded) are trained on \dataset from scratch, keeping only generic pretrained
encoders, or fine-tuned from the pretrained single-region models as in
\cref{tab:benchmark}. Results are averaged over three seeds, with a standard deviation below 0.5 on every metric.}
\label{tab:benchmark-scratch}
\small
\setlength{\tabcolsep}{2pt}
\begin{tabularx}{\linewidth}{l*{4}{>{\centering\arraybackslash}X}c}
\toprule
Method & aIoU\,$\uparrow$ & AUC\,$\uparrow$ & SIM\,$\uparrow$ & MAE\,$\downarrow$ & Role\,$\uparrow$ \\
\midrule
\multicolumn{6}{l}{OpenAD~\cite{nguyen2023openad}} \\
\quad Zero-shot & 3.9 & 50.4 & 37.9 & 0.332 & -- \\
\rowcolor{gray!12}
\quad From scratch & 26.1 & 70.3 & 53.8 & 0.281 & 73.1 \\
\rowcolor{gray!12}
\quad Fine-tuned & 25.4 & 70.4 & 53.7 & 0.275 & 73.1 \\
\midrule
\multicolumn{6}{l}{PointRefer~\cite{li2024laso}} \\
\quad Zero-shot & 7.3 & 51.3 & 32.7 & 0.332 & -- \\
\rowcolor{gray!12}
\quad From scratch & 36.4 & 79.1 & 62.8 & 0.190 & 80.1 \\
\rowcolor{gray!12}
\quad Fine-tuned & 34.8 & 78.6 & 61.8 & 0.198 & 79.1 \\
\midrule
\multicolumn{6}{l}{Espresso-3D~\cite{lee2025affogato}} \\
\quad Zero-shot & 10.3 & 53.7 & 39.8 & 0.331 & -- \\
\rowcolor{gray!12}
\quad From scratch & 37.4 & 79.3 & 62.5 & 0.192 & 78.8 \\
\rowcolor{gray!12}
\quad Fine-tuned & 37.8 & 79.5 & 62.8 & 0.191 & 78.8 \\
\bottomrule
\end{tabularx}
\end{table}

\paragraph{Overlap between the two role heatmaps.}
To measure whether a model separates the two roles of a task, we compute the
IoU between the 10\% highest-scoring points of its two heatmaps, averaged over
the 9{,}502 test tasks with distinct roles. \cref{tab:role-overlap} shows that
the annotated contacts of the two roles barely overlap, whereas each zero-shot
model places its two heatmaps largely on the same points, although its
instructions name both roles. After training on \dataset, the two heatmaps of
every model become as disjoint as the annotated ones. They remain disjoint on
the tasks with identical roles, with an overlap of at most 0.02.

\begin{table}[t]
\centering
\caption{\textbf{Overlap between the two role heatmaps of a task.} IoU between
the 10\% highest-scoring points of the two heatmaps, averaged over the
9{,}502 test tasks with distinct roles. Zero-shot models are queried once per
role. Dual models are averaged over three random seeds.}
\label{tab:role-overlap}
\small
\begin{tabular}{@{}lc@{}}
\toprule
Heatmaps & Overlap \\
\midrule
Annotated & 0.01 \\
\midrule
OpenAD~\cite{nguyen2023openad}, zero-shot & 0.52 \\
PointRefer~\cite{li2024laso}, zero-shot & 0.56 \\
Espresso-3D~\cite{lee2025affogato}, zero-shot & 0.71 \\
\midrule
Dual-OpenAD & 0.00 \\
Dual-PointRefer & 0.00 \\
Dual-Espresso-3D & 0.01 \\
\bottomrule
\end{tabular}
\end{table}

\section{Bimanual Manipulation on Unseen Objects}
\label{sec:unseen-results}

\begin{figure*}[t]
\centering
\includegraphics[width=\textwidth]{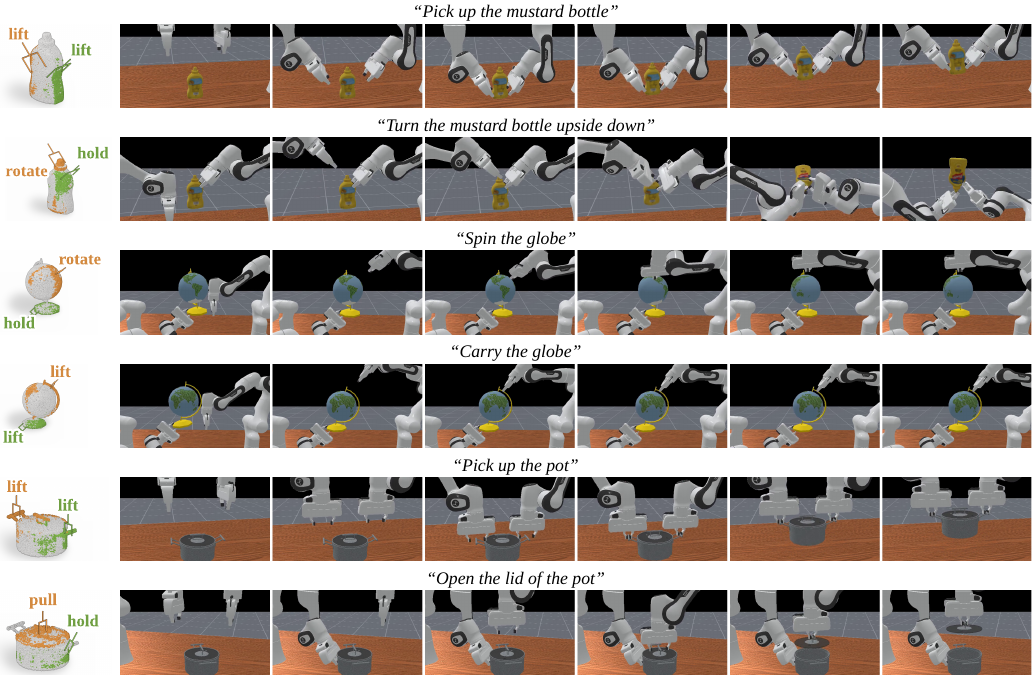}
\caption{\textbf{More bimanual manipulation on unseen objects.} Each object, from YCB or PhysX-Mobility, appears with two tasks, one per row. The left panel shows the top 10\% highest-scoring heatmap regions, the grasps used in the rollout, and the predicted role verbs. The frames on the right show the manipulation sequence. The predicted regions and roles change with the task on the same object.}
\label{fig:sim-more}
\end{figure*}

\subsection{Simulation Setup}
\label{sec:sim-setup}

\dataset objects contain only visual geometry, without the mass, collision meshes, or joints that physics simulation requires.
We therefore select objects from three sources that provide physical assets, where YCB~\cite{calli2015ycb} and OakInk2~\cite{zhan2024oakink2} offer scanned real-world objects and PhysX-Mobility~\cite{cao2026physxanything} offers articulated objects with physical annotations.
From these sources, we collect 30 objects that suit bimanual manipulation with two parallel-jaw grippers, including 13 from YCB, 13 from OakInk2, and 4 from PhysX-Mobility.
Each object is paired with two distinct tasks, such as ``open the laptop'' and ``carry the laptop'', yielding 60 tasks.
For each task, Dual-Espresso-3D receives a 16{,}384-point cloud and the task query, and returns one heatmap and one role verb per hand.

We run all rollouts in ManiSkill3~\cite{tao2025maniskill3} with CPU physics at 100\,Hz and a control rate of 20\,Hz.
Two Franka Panda arms face each other across a table, 1.1\,m apart, and each arm follows end-effector pose targets with a binary gripper command.
Each arm grasps the region formed by the 10\% highest-scoring points of its role heatmap, with a top-down or edge pinch that fits the gripper.
Scripted motions then execute each task according to the predicted role verbs, verifying that grasps placed in the predicted affordance regions can accomplish the task.

\subsection{Additional Results}
\label{sec:sim-more}
\cref{fig:sim-more} visualizes predicted heatmaps and simulated rollouts on three of these unseen objects.
We evaluate Dual-Espresso-3D on all 60 tasks by visually inspecting whether the 10\% highest-scoring points of each role lie on a contact region suited to the task.
The predicted regions are suitable for 55 of the 60 tasks.

The predictions adapt to the task on the same object.
Picking up the mustard bottle places the two hands on opposite narrow sides, whereas turning it upside down moves them to the cap and the upper body.
Spinning the globe targets the sphere and the base, while carrying it targets the meridian arc and the base.
Picking up the pot targets its two side handles, while opening the lid targets the lid and the lower wall.
The predicted role verbs follow the task as well.
Tasks that move the whole object assign lift to both hands, while tasks that change its pose or state pair an active role, such as rotate or pull, with hold.
On the articulated globe and pot, the active role falls on the movable part, the sphere or the lid, and hold falls on the stationary part.

We execute representative tasks with the setup in \cref{sec:sim-setup}.
In each rollout, the two arms grasp the predicted regions and complete the task, showing that the predicted affordances guide bimanual manipulation on unseen objects.

\section{Source Assets and Release}
\label{sec:source-assets}

\paragraph{Relation between the sources.}
The three sources in \cref{sec:source-assets-main} describe the same objects.
G-Objaverse~\cite{zuo2024sparse3d} renders Objaverse~\cite{deitke2023objaverse}
assets from fixed cameras, and Affogato~\cite{lee2025affogato} provides point
clouds of the same assets. We match the two through the Objaverse object
identifier, so every object in \dataset has its 40 RGB-D views with camera
parameters and a point cloud of the same asset. The views and the point cloud
share the normalized object frame and differ only by a fixed change of axis
convention, which we apply to every point cloud. Stage~4 can therefore project
the 2D masks onto the point cloud with the G-Objaverse cameras and depth maps,
without per-object registration. Building on the Affogato point clouds also keeps
\dataset compatible with models and benchmarks developed on Affogato, since the
same points carry both annotations.

\paragraph{Licenses and release.}
Objaverse is distributed under the ODC-By~1.0 license, and each of its objects
carries a Creative Commons license recorded in its metadata. G-Objaverse is
released under the MIT license, and the Affogato point clouds are publicly
available. We will release the \dataset annotations, i.e., the tasks, role
pairs, and heatmaps, indexed by the identifiers of their source objects, under the
CC~BY~4.0 license, together with the code of the annotation pipeline and the
benchmark under the MIT license. Each object remains
subject to the license of its original asset.
Because every stage runs on open-weight models at under one GPU-minute per
object, the pipeline can be reproduced and extended to new objects at low cost.

\section{Limitations and Future Work}
\label{sec:limitations}

\dataset annotates where two hands make contact on an object and which role each hand plays in a task. A natural next step is to describe how each role is carried out. Enriching the roles with force direction and magnitude, approach direction, and grasp type would let models also predict the physical parameters of each hand's action. Contacts in everyday manipulation also depend on the surroundings and on the preceding and following actions. This motivates extending the annotations from a single object and role pair to scene context, object state, and multi-step tasks. On the benchmark side, models currently receive complete object point clouds, whereas a robot observes objects from its own viewpoint. Extending the benchmark to partial point clouds is therefore a promising direction. The broad coverage of everyday objects and tasks in \dataset provides prior knowledge that such models can transfer to these settings.

\section{Annotation and Judge Prompts}
\label{sec:annotation-prompts}

\cref{fig:prompt-stage1,fig:prompt-stage2-tasks,fig:prompt-stage2-roles} show the prompts that Gemma~4 receives in Stages~1 and~2 of the pipeline, condensed for space. Each prompt also receives the semantic views $\mathcal V_i^{\mathrm{sem}}$, and Gemma~4 decodes greedily into a fixed JSON format. \cref{fig:judge-prompt-task,fig:judge-prompt-role,fig:judge-prompt-heatmap} show the prompts of the judge (\cref{sec:quality-protocol}).

\begin{figure*}[p]
\centering
\begin{promptbox}{Stage 1 prompt: object filtering}
\promptlabel{System}You are a data quality filter for a bimanual (two-arm) robot manipulation dataset. Decide whether a 3D object should be included. Keep only real, everyday functional objects that people handle with their hands and that have at least one realistic task requiring both hands to coordinate at the same time. When in doubt, reject. Judge from the rendered 3D geometry, not the category name.
\tcbline
\promptlabel{User}{\ttfamily\scriptsize \textit{views, including a top-down and an underside view}\\Object category: $\langle$object name$\rangle$}\par
Keep the object only if both hold:\par
1. Everyday functional object: a real, common object that people pick up and use with their hands. Reject an object that is mainly decorative or that people do not routinely handle in order to use it.\par
2. Bimanual task exists: at least one realistic task in which two hands coordinate to operate or move the object, such as carrying a large object, opening, folding, twisting, or stabilizing one part while actuating another.
\tcbline
\promptlabel{Response}{\ttfamily\scriptsize \{"everyday\_object": true $|$ false, "bimanual\_task\_exists": true $|$ false, "example\_task": $\langle$text$\rangle$ $|$ null, "keep": true $|$ false\}}
\end{promptbox}

\vspace{6pt}
\begin{promptbox}{Stage 1 prompt: component-usage curation}
\promptlabel{System}You are a visual parser specializing in object manipulation analysis.
\tcbline
\promptlabel{User}{\ttfamily\scriptsize \textit{views, including a top-down and an underside view}\\The object category is: $\langle$object name$\rangle$}\par
Identify all components that a human hand can manipulate during everyday use, using the top-down and underside views to find parts that are not visible from the side. Be thorough and fine-grained: an object usually has several manipulable parts. Always include the rigid body itself, then every other part that a hand uses, such as an operable part or a visible grip feature.\par
Rules:\par
\hspace*{1.2em}List only parts visible in these views, not parts inferred from the category name.\par
\hspace*{1.2em}List a part that opens or moves only if a visible feature shows that it operates, such as a seam, a hinge, a thread, or a graspable edge.\par
\hspace*{1.2em}Exclude decorative parts such as logos, text, and labels.\par
\hspace*{1.2em}Give each part a short name and a one-sentence description of its interaction, and merge parts that serve the same purpose.
\tcbline
\promptlabel{Response}{\ttfamily\scriptsize \{"canonical\_components": [\{"name": $\langle$text$\rangle$, "interaction": $\langle$text$\rangle$\}, \ldots]\}}
\end{promptbox}
\caption{\textbf{Stage~1 prompts (condensed).} The first prompt decides whether an object enters \dataset; the second extracts the component-usage pairs $\mathcal C_i$ of a kept object.}
\label{fig:prompt-stage1}
\end{figure*}

\begin{figure*}[p]
\centering
\begin{promptbox}{Stage 2 prompt: multi-granular task proposal}
\promptlabel{System}You are an annotation assistant for a 3D bimanual affordance dataset. Propose daily bimanual manipulation tasks for one object. A valid task is a normal user operation that the object is designed to support, without damage or tools. Do not generate filler tasks; returning empty lists is correct when no good task exists.
\tcbline
\promptlabel{User}{\ttfamily\scriptsize \textit{views}\\Object: $\langle$object name$\rangle$\\Groundable components: $\langle$component-usage pairs$\rangle$\\Already proposed tasks to avoid repeating: $\langle$tasks$\rangle$}\par
Task families:\par
\hspace*{1.2em}A) object\_level: the object is manipulated mainly as a whole object.\par
\hspace*{1.2em}B) component\_level: existing components change state, configuration, access, attachment, or relation to another part of the object. The component must be visible, listed, or clearly designed to move or separate.\par
\hspace*{1.2em}C) functional\_pose: the object is put into a use pose that enables an external goal, such as level, upright, tilted, aimed, spread, taut, opened, or aligned for use, without performing the external action.\par
Think like a person using the object normally. Look for designed operations that the object visibly supports, such as open and close, lock and unlock, fold and unfold, roll and unroll, extend and retract, tighten and loosen, adjusting or rotating a built-in part, operating a handle, knob, button, or latch, removing or replacing a removable part, or pouring contents already held by the object. For functional\_pose, prefer diverse use poses, such as receiving, dispensing, aiming, keeping a surface level, keeping an opening accessible, or keeping a flexible body taut.\par
Each task needs a concrete goal written as ``$\langle$before$\rangle$ $\rightarrow$ $\langle$after$\rangle$'' that changes the object itself, one of its components, or contents it already holds; a change in how the hands hold the object is not a goal. Prefer component\_level tasks when meaningful components exist. If the object has several designed operations, include several distinct tasks; if it has only one or two real operations, return only those.
\tcbline
\promptlabel{Response}{\ttfamily\scriptsize \{"object\_level": [\{"task": $\langle$text$\rangle$, "goal": $\langle$before$\rangle$ $\rightarrow$ $\langle$after$\rangle$\}, \ldots], "component\_level": [\ldots], "functional\_pose": [\ldots]\}}
\end{promptbox}

\vspace{6pt}
\begin{promptbox}{Stage 2 prompt: task curation}
\promptlabel{System}You are a strict curator for a 3D bimanual affordance dataset. Keep only realistic daily bimanual tasks. A valid task is a normal user operation that the object is designed to support, without damage or tools. When uncertain, reject.
\tcbline
\promptlabel{User}{\ttfamily\scriptsize \textit{views}\\Object: $\langle$object name$\rangle$\\Groundable components: $\langle$component-usage pairs$\rangle$\\Candidate tasks: $\langle$proposed tasks of each family$\rangle$}\par
Compose the final task list. Keep a candidate only if:\par
\hspace*{1.2em}its goal is a concrete ``$\langle$before$\rangle$ $\rightarrow$ $\langle$after$\rangle$'' change verifiable from the object itself;\par
\hspace*{1.2em}it is a normal operation an ordinary person would intentionally do to this object;\par
\hspace*{1.2em}the object visibly supports the change without damage;\par
\hspace*{1.2em}it naturally requires or strongly benefits from two hands;\par
\hspace*{1.2em}it is not already covered by another kept task.\par
For functional\_pose, keep diverse pose types instead of many variants of one pose. Rewrite kept tasks into simple daily task phrases.
\tcbline
\promptlabel{Response}{\ttfamily\scriptsize \{"tasks": [\{"task": $\langle$text$\rangle$, "goal": $\langle$before$\rangle$ $\rightarrow$ $\langle$after$\rangle$, "why\_bimanual": $\langle$text$\rangle$, "category": object\_level $|$ component\_level $|$ functional\_pose\}, \ldots]\}}
\end{promptbox}
\caption{\textbf{Stage~2 prompts for task generation (condensed).} The first prompt proposes candidate tasks over several passes, each listing the tasks already proposed, and the second keeps only valid and distinct ones. The families object\_level, component\_level, and functional\_pose are the object-, component-, and function-level families of \cref{tab:task-families}.}
\label{fig:prompt-stage2-tasks}
\end{figure*}

\begin{figure*}[p]
\centering
\begin{promptbox}{Stage 2 prompt: operation mechanics}
\promptlabel{System}You are a manipulation-mechanics expert. For one task on one object, work out the physics of performing it: which part moves, which way it must go, where to grip it for leverage, and what the other hand must hold.
\tcbline
\promptlabel{User}{\ttfamily\scriptsize \textit{views}\\Object: $\langle$object name$\rangle$\quad Task: $\langle$task$\rangle$\quad Goal: $\langle$goal$\rangle$\\Visible parts: $\langle$components$\rangle$\\Allowed verbs: $\langle$eight verbs and their meanings$\rangle$}\par
1. Mechanism and scale: the moving element, i.e., what the goal moves (the whole body, a part, or the contents), and the stable frame that stays fixed as the reference. Judge the real-world size class from what the object is, since the views are size-normalized.\par
2. Operation mechanics, if a part moves: its allowed motion (hinge swing, slide, or twist) and where the hinge, track, or axis is; the direction from the current to the goal state, with an acting verb that produces force in that direction; the purchase, a rigid contact farthest from the hinge or axis; and the reaction, a different part that the other hand holds stationary while the acting part moves.\par
3. Coordination: stabilize+actuate when one part moves on a fixed body, co-actuate when both hands drive moving parts at once, and whole-body when nothing on the object moves.
\tcbline
\promptlabel{Response}{\ttfamily\scriptsize \{"size\_class", "moving\_part", "motion", "direction\_to\_goal", "acting\_verb", "purchase\_contact", "reaction\_part", "coordination": $\langle$text$\rangle$\}}
\end{promptbox}

\vspace{6pt}
\begin{promptbox}{Stage 2 prompt: role pair}
\promptlabel{System}You convert the operation mechanics into two coordinated, hand-agnostic robot roles. The physics is already worked out; transcribe it into roles.
\tcbline
\promptlabel{User}{\ttfamily\scriptsize \textit{views}\\Object: $\langle$object name$\rangle$\quad Task: $\langle$task$\rangle$\quad Goal: $\langle$goal$\rangle$\\Visible parts: $\langle$components$\rangle$\\Allowed verbs: $\langle$eight verbs$\rangle$\\Operation mechanics: $\langle$mechanics$\rangle$}\par
Make exactly two roles, A and B. For stabilize+actuate, one hand applies the acting verb at the purchase contact and the other holds the reaction part. For co-actuate, each hand drives its own moving part, or the two hands pull two halves apart. For a whole-body lift, a hand on a smooth side face bears weight only as half of a two-hand squeeze, so both hands lift opposite faces or two handles together; when a single dominant grip, such as a mug's handle, can carry a light object, that hand lifts the grip and the other steadies the body.\par
Each role has a verb, a target part from the visible parts, a contact region on that part, and a function. The two contact regions are different places, reachable in the object's natural resting pose, and pinned in the object's own frame so that the same spot can be found from any view: the up-axis (top or bottom, rim or base), a named part, or a structural feature such as a seam, hinge, or corner. Never use a viewer direction such as left or right, which changes with the camera.\par
Both hands doing the same action on the same spot is not bimanual; symmetric roles act at different locations, such as opposite faces in a co-lift. Write the relation as the combined effect of the two roles toward the goal.
\tcbline
\promptlabel{Response}{\ttfamily\scriptsize \{"roles": [\{"id": "A", "role": $\langle$verb$\rangle$, "target": $\langle$part$\rangle$, "contact\_region": $\langle$text$\rangle$, "function": $\langle$text$\rangle$\}, \{"id": "B", \ldots\}], "relation": $\langle$text$\rangle$\}}
\end{promptbox}
\caption{\textbf{Stage~2 prompts for role decomposition (condensed).} The operation plan of \cref{sec:pipeline} is derived in two steps: the first prompt works out the mechanics of a task, and the second turns them into the role pair $r_{i,k}$. The contact region and function of each role fill its pointing query in Stage~2.}
\label{fig:prompt-stage2-roles}
\end{figure*}

\begin{figure*}[p]
\centering
\begin{promptbox}{Judge prompt: task validity}
\promptlabel{System}You will see one object in eight views and a task someone proposes to do with it using two hands. Using what you see and your everyday knowledge of objects like this, judge whether the task makes sense for this object and is a meaningful thing to do with it. The object name can be wrong, so go by what the views show.\par
\hspace*{1.2em}2 = it makes sense and is meaningful for this object.\par
\hspace*{1.2em}1 = unclear: you cannot tell whether it makes sense, or whether it has a real point.\par
\hspace*{1.2em}0 = it clearly does not make sense or has no real point for this object.\par
First say whether the parts the task needs are visible in the views (yes, no or unsure), then give the score and one short sentence explaining it.
\tcbline
\promptlabel{User}{\ttfamily\scriptsize \textit{eight views, each after the caption} View: $\langle$label$\rangle$\\Object: $\langle$object name$\rangle$\\ Task: $\langle$task$\rangle$}
\tcbline
\promptlabel{Response}{\ttfamily\scriptsize \{"parts\_visible": yes $|$ no $|$ unsure, "score": 0 $|$ 1 $|$ 2, "reason": $\langle$text$\rangle$\}}
\end{promptbox}
\caption{\textbf{Judge prompt for task validity (condensed).} The system prompt, the user input, and the response format of the judge. The judge sees eight views of the object and the task, and returns a score from 0 (invalid) to 2 (valid) with a one-sentence justification.}
\label{fig:judge-prompt-task}
\end{figure*}

\begin{figure*}[p]
\centering
\begin{promptbox}{Judge prompt: role validity}
\promptlabel{System}You check affordance annotations for 3D object models. An annotation tells a person or robot with two hands how to act on the object to do a task: for each hand, an action, the part it touches, where on that part, and what it is for. The touched part and region are where the hand makes contact and applies its action.\par
The action is chosen from eight verbs (hold, lift, push, pull, press, slide, rotate, squeeze).\par
You get eight views of the object (they show only the object, not the hands), its name, the task, and the two roles.\par
Answer in these steps:\par
1. plan: in one sentence, how a person would normally do this task with two hands.\par
2. parts: are the parts the hands touch on this object? (yes, no or unsure)\par
3. score:\par
\hspace*{1.2em}2 = the roles are a reasonable way to do the task with these eight verbs.\par
\hspace*{1.2em}1 = you cannot tell.\par
\hspace*{1.2em}0 = clearly not a way to do the task.\par
4. reason: one short sentence.\par
Judge the roles for the task as given.
\tcbline
\promptlabel{User}{\ttfamily\scriptsize \textit{eight views, each after the caption} View: $\langle$label$\rangle$\\Object: $\langle$object name$\rangle$\\ Task: $\langle$task$\rangle$\\Hand A: touches $\langle$target$\rangle$\ at $\langle$contact region$\rangle$; action: $\langle$role$\rangle$; purpose: $\langle$function$\rangle$\\Hand B: \textit{as for Hand A}}
\tcbline
\promptlabel{Response}{\ttfamily\scriptsize \{"plan": $\langle$text$\rangle$, "parts": yes $|$ no $|$ unsure, "score": 0 $|$ 1 $|$ 2, "reason": $\langle$text$\rangle$\}}
\end{promptbox}
\caption{\textbf{Judge prompt for role validity (condensed).} The input adds the role pair, with Hand~A and Hand~B denoting $r^a$ and $r^b$. The judge first describes how a person would perform the task, then scores the role pair.}
\label{fig:judge-prompt-role}
\end{figure*}

\begin{figure*}[p]
\centering
\begin{promptbox}{Judge prompt: heatmap quality}
\promptlabel{System}You check contact heatmaps in affordance annotations for 3D object models. An annotation tells a person or robot with two hands how to act on an object to do a task. For each hand there is a role (the part it touches, the action and its purpose) and a heatmap.\par
What a heatmap shows: all the possible places on the object where that hand could make contact to do its role. The object is drawn as dark points and the region is colored, orange for Hand A and teal for Hand B; brighter means a more likely contact. A usable region is one clear area, or a few, on places where the hand could do its job; specks scattered over unrelated parts, or a trace too faint to see, give the hand no clear place to act.\par
The size of a region does not by itself make it good or bad: a hand that holds, supports or moves the whole object can touch it in many places, while a hand that works one particular part can only touch that part.\par
You get three images of the object from the same eight cameras: the object itself, Hand A's heatmap and Hand B's heatmap. The role text says roughly where each hand acts; judge the regions you see against the task.\par
Answer in these steps:\par
1. hand\_a\_region: look at the orange points and say where they are on the object and how they look: one clear area, a few separate areas, scattered specks, a faint trace, or spread over most of the object.\par
2. hand\_a\_quality: is the orange region clean (on places where Hand A can do its action, with nothing notable elsewhere), flawed (usable, but part of it lies elsewhere, it is broken into small bits, or it is very small) or unusable (missing, too faint, or not where Hand A can do its action)?\par
3. hand\_b\_region: the same as step 1 for the teal points.\par
4. hand\_b\_quality: the same as step 2 for Hand B.\par
5. score:\par
\hspace*{1.2em}2 = both regions are clean.\par
\hspace*{1.2em}1 = a region is flawed.\par
\hspace*{1.2em}0 = a region is unusable.\par
6. reason: one short sentence.
\tcbline
\promptlabel{User}{\ttfamily\scriptsize Object views: \textit{[image]} \quad Hand A heatmap (orange): \textit{[image]} \quad Hand B heatmap (teal): \textit{[image]}\\Object: $\langle$object name$\rangle$\\ Task: $\langle$task$\rangle$\\Hand A: touches $\langle$target$\rangle$; action: $\langle$role$\rangle$; purpose: $\langle$function$\rangle$\\Hand B: \textit{as for Hand A}}
\tcbline
\promptlabel{Response}{\ttfamily\scriptsize \{"hand\_a\_region": $\langle$text$\rangle$, "hand\_a\_quality": clean $|$ flawed $|$ unusable, "hand\_b\_region": $\langle$text$\rangle$, "hand\_b\_quality": clean $|$ flawed $|$ unusable, "score": 0 $|$ 1 $|$ 2, "reason": $\langle$text$\rangle$\}}
\end{promptbox}
\caption{\textbf{Judge prompt for heatmap quality (condensed).} The input shows the object and the heatmaps of $r^a$ (orange) and $r^b$ (teal), rendered from the same eight cameras. The judge describes and grades each region before scoring the heatmap pair.}
\label{fig:judge-prompt-heatmap}
\end{figure*}

\end{document}